%% file: acl_latex.tex
\documentclass[11pt]{article}

\usepackage[final]{acl}

\usepackage{times}
\usepackage{latexsym}

\usepackage[T1]{fontenc}
\usepackage[utf8]{inputenc}

\usepackage{microtype}

\usepackage{inconsolata}

\usepackage{hyperref}
\usepackage{url}
\usepackage{graphicx}
\usepackage{booktabs}       % professional-quality tables
\usepackage{amsfonts}       % blackboard math symbols
\usepackage{nicefrac}       % compact symbols for 1/2, etc.
\usepackage{amsmath}
\usepackage{microtype}      % microtypography
\usepackage{xcolor}         % colors
\usepackage{xpatch}
\usepackage{booktabs}
\usepackage{xpatch}
\usepackage[table]{xcolor}
\usepackage{wrapfig}
\usepackage{amsfonts} 
\usepackage{subcaption}
\usepackage{algorithm}
\usepackage{bbding}
\usepackage{algpseudocode}
\usepackage{multirow}
\usepackage[table]{xcolor}
\usepackage{multicol}
\usepackage{caption}
\usepackage{subcaption}
\usepackage{enumitem}
\usepackage{placeins} 
\usepackage{tcolorbox}
\tcbuselibrary{theorems, skins} % 必须加载 theorems 库
\newtcbtheorem{lem}{}{%
    colback=orange!5,
    colframe=orange!65,
    fonttitle=\bfseries,
    detach title,
    boxsep=1pt,
    left=5pt,
    right=5pt,
    top=5pt,
    bottom=5pt,
}{lem}
    
\title{Does Theory of Mind Improvement Really Benefit Human-AI Interactions? Empirical Findings from Interactive Evaluations 
}
\newcommand{\revise}[1]{\textcolor{black}{#1}}

\author{
  \textbf{Nanxu Gong\textsuperscript{1}$^*$},
  \textbf{Zixin Chen\textsuperscript{2}$^*$},
  \textbf{Haotian Li\textsuperscript{3}$^\dagger$},
  \textbf{Zishu Zhao\textsuperscript{4}$^*$},
\\
  \textbf{Jianxun Lian\textsuperscript{3}},
  \textbf{Huamin Qu\textsuperscript{2}},
  \textbf{Yanjie Fu\textsuperscript{1}},
  \textbf{Xing Xie\textsuperscript{3}}
\\
\\
  \textsuperscript{1}Arizona State University,
  \textsuperscript{2}Hong Kong University of Science and Technology,\\
  \textsuperscript{3}Microsoft Research Asia,
  \textsuperscript{4}Smith College
\\
\small{
  \textbf{Correspondence:} \href{mailto:haotian.li@microsoft.com}{haotian.li@microsoft.com}, \href{mailto:nanxugong@outlook.com}{nanxugong@outlook.com}
}
}

\begin{document}
\maketitle
\renewcommand{\thefootnote}{\fnsymbol{footnote}}

\footnotetext[1]{Work done during internship at Microsoft Research Asia.}
\footnotetext[2]{Corresponding author.} 
\input{Abstract_v2}
\input{Introduction_v2}
\input{Problem}
\input{Method}
\input{Experiment}

\input{Related}
\input{Conclusion}

\bibliography{Reference}

\appendix

\input{Appendix}

\end{document}

%% file: Abstract_v2.tex
\begin{abstract}
\begin{figure*}[t]
    \centering
    % Use resizebox to scale the image
    \resizebox{\textwidth}{!}{%
      \includegraphics{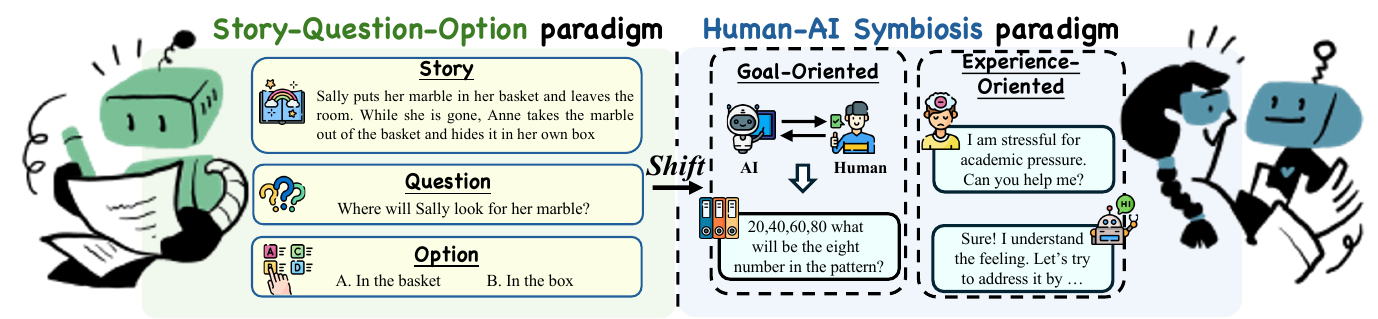}%
    }
    \caption{Based on a new dynamic and interactive evaluation paradigm, our research explores the effectiveness of LLMs with existing ToM enhancement techniques for HAI symbiosis.}
    \label{fig:problem}
    \vspace{-0.4cm}
\end{figure*}

Improving the Theory of Mind (ToM) capability of Large Language Models (LLMs) is crucial for effective social interactions between these AI models and humans.
However, the existing benchmarks often measure ToM capability improvement through story-reading, multiple-choice questions from a third-person perspective, while ignoring the first-person, dynamic, and open-ended nature of human-AI (HAI) interactions.
To directly examine how ToM improvement techniques benefit HAI interactions, we first proposed the new paradigm of interactive ToM evaluation with both perspective and metric shifts.
Next, following the paradigm, we conducted a systematic study of four representative ToM enhancement techniques using both four 
\revise{benchmarks based on real-world scenarios}
% real-world datasets 
and a user study, covering both goal-oriented tasks (e.g., coding, math) and experience-oriented tasks (e.g., counseling).   
Our findings reveal that improvements on static benchmarks do not always translate to better performance in dynamic HAI interactions. This paper offers critical insights into ToM evaluation,  showing the necessity of interaction-based assessments in developing next-generation, socially aware LLMs for HAI symbiosis. 

\end{abstract}

%% file: Introduction_v2.tex
\section{Introduction}

Theory of Mind (ToM) denotes the cognitive capacity to attribute unobservable mental states (e.g., beliefs, intentions, emotions), which is essential for social interaction \citep{survey_assess_enhance, 
survey1, strachan2024testing}. 
As a foundational component of social cognition, ToM has been recognized as a core social intelligence skill that advanced LLMs should obsess to improve their interactions with humans and ultimately achieve the target of human-AI symbiosis~\cite{street2024llm}.

To nurture LLMs' ToM capability, the cornerstone is to understand its capability levels and benefits led by improvement methods with appropriate and sufficient evaluation.
To achieve the goal, the existing dominant methods are
% However, current ToM evaluation methods are dominated by 
static, task-based assessments in a story-question-option format, an approach derived from classic false-belief tests like the Sally-Anne task, such as the one by Kosinski~\cite{kosinski2024evaluating}. 
% However, one critical concern is the risk of memorizing the results of these simple tasks in the training process of LLMs.
Following the design, subsequent benchmarks such as HiToM \citep{HiToM} and ToMBench \citep{ToMBench} have increased the complexity and diversity of these tests.
More recent benchmarks, such as ExploreToM~\cite{ExploreToM}, leverage adversarial methods to further increase the diversity of problems and reduce the risk of memorizing benchmarks in the LLM training process.
However, these third-person, story-reading benchmarks with accuracy as the only standard fail to ground ToM evaluation in the real-world context of Human-AI (HAI) interaction and collaboration.
In HAI scenarios, LLMs are supposed to leverage their ToM capability to perform \textbf{first-person perspective} actions in response to \textbf{dynamic} and sometimes \textbf{open-ended} user requests with \textbf{diverse} targeted metrics.
% with their understanding of users' beliefs and intents.
The mismatch of task natures creates a critical \textit{socio-technical gap}~\cite{liao2023rethinking} between benchmark performance and real-world competence, where \textbf{the ToM benchmark result improvements may not lead to sensible benefits for human-LLM interactions.}

To reveal the gap and guide future ToM evaluation, our research examines how these benchmark improvements are transformed into real-world values through a new paradigm of interactive ToM evaluation.
% we introduce an application-grounded approach that aligns evaluation with actual socio-requirements. 
We first shift the ToM task from a static, third-person perspective into a dynamic, open-ended, and first-person one, where the LLM agent engages in multi-turn conversations across diverse and real-world scenarios. 
% This allows us to assess model's practical ability to satisfy human needs in deployment contexts with both perspective and evaluation metrics shift. 
Next, drawing from cognitive science \citep{goal-experience1,goal-experience2,bales1950interaction}, we classify these scenarios into two primary categories based on the evaluation objective: goal-oriented tasks (e.g., math, code) and experience-oriented tasks (e.g., counseling, healthcare).
Then, we simulate real-world HAI interactions in nine tasks under the two types of scenarios and leverage task-specific metrics (e.g., accuracy, success rate) to evaluate the performance of LLMs with ToM enhancement techniques, including both prompt-based techniques and finetuning-based ones. 
By aggregating the results from four benchmarks, we comprehensively assess the effectiveness of ToM enhancement techniques across nine domains that align well with actual user requirements. 
Furthermore, we also conduct a crowdsourcing user study to support our findings, ensuring the results reflect genuine human perceptions.
Our rigorous evaluation reveals three key insights regarding current ToM enhancement techniques: \textbf{(i) A Performance Gap in Evaluation}: There is a significant gap between how models perform on static, story-based ToM benchmarks and their actual capabilities in dynamic, interactive scenarios, showing that current evaluation methods are insufficient for measuring readiness for HAI collaboration. \textbf{(ii) A Failure to Generalize}: ToM enhancement techniques improve a model’s performance in experience-oriented tasks but fail to generalize this success to goal-oriented tasks, separating the capability requirements in various real-world scenarios.
\textbf{(iii) A Gap in User Perception}: The modest gains from current ToM methods are often too subtle to cross a user's perceptual threshold, meaning the improvements measured in benchmarks do not translate into a meaningfully better user experience.
% \haotian{Also, summarize our key insights here.}
Our contributions include: 
% \vspace{-0.8cm}

\begin{itemize}[leftmargin=10pt, nosep]
\item We shift ToM evaluation from static tests to dynamic, real-world HAI interactions. \item We assess ToM enhancement methods in goal- and experience-oriented scenarios via simulated interactive benchmarks and user studies. \item We identify critical limitations in current ToM enhancement methods and provide insights for future research.
\end{itemize}

%% file: Problem.tex
\section{Interactive ToM Evaluation Paradigm}

\subsection{Background: Existing ToM Evaluation Paradigm with Static Benchmarks}

ToM evaluation in existing benchmarks is typically operationalized through a static, story-question-option format. 
Formally, given a story $S = \{s_1, s_2, \dots, s_n\}$ and a question $Q$, the model must select the correct answer from a candidate set $O = \{o_1, \dots, o_k\}$, where only one option $o_{\text{correct}}$ is correct:
\begin{equation}
o^* = \arg\max_{o_i \in O} P(o_i \mid S, Q).
\end{equation}
Performance is then measured by accuracy:
\begin{equation}
\text{Acc} = \tfrac{1}{N} \sum_{i=1}^{N} \mathbb{I}(o_i^* = o_{i,\text{correct}}),
\end{equation}
where $N$ is the number of test samples.  
This formulation captures a \emph{static evaluation paradigm}, where reasoning occurs over a fixed textual world.
It is hard to reflect the open-ended, dynamic, and multi-turn nature of human–AI interactions, where responses are not unique and their satisfactory levels cannot be simply judged as binary outcomes.
% \haotian{Please check the last sentence here.}

\begin{figure*}[t]
    \centering
    % Use resizebox to scale the image
    \resizebox{\textwidth}{!}{%
      \includegraphics{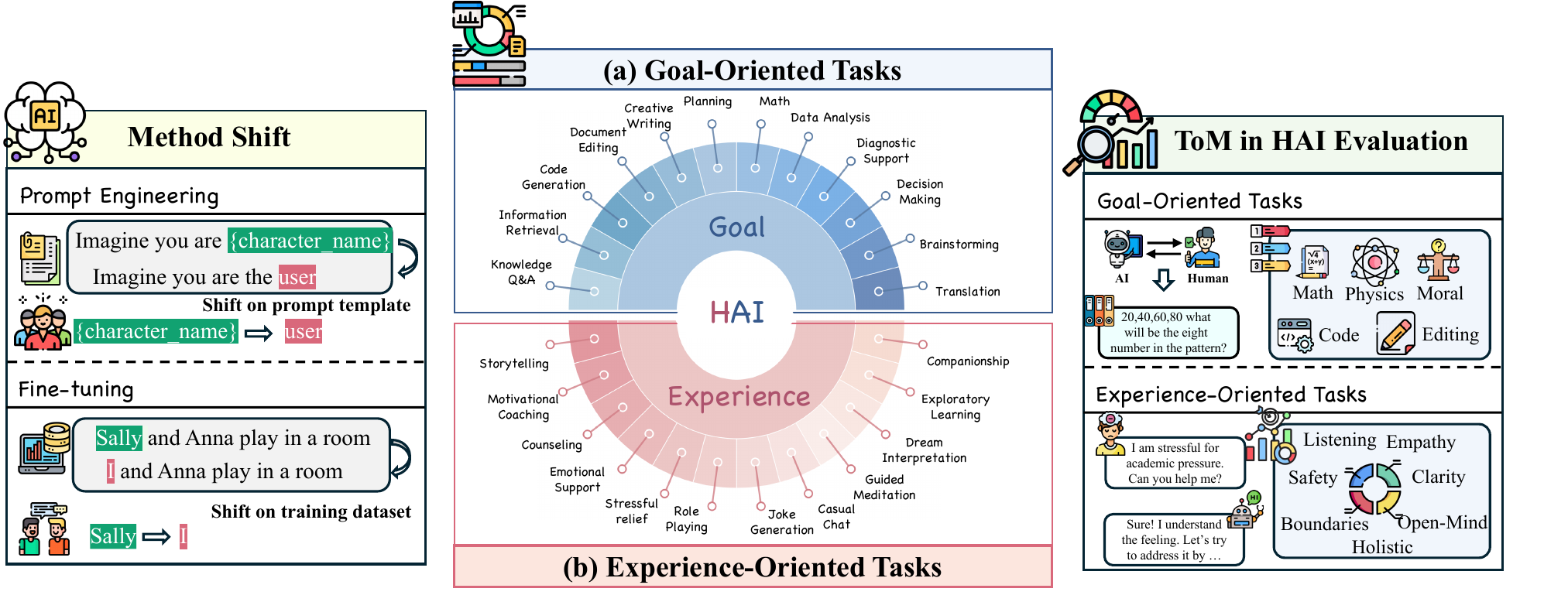}%
    }
    \caption{Overview of our interactive ToM evaluation paradigm for real-world HAI interaction.}
    \label{fig:framework}
    \vspace{-0.4cm}
\end{figure*}

\subsection{Our Paradigm: Pivot ToM Evaluation to Interactive HAI Settings}
A large body of developmental, longitudinal, and neurocognitive work indicates that stronger ToM is associated with 
richer social competence, more cooperative behaviors, and more effective joint action~\citep{imuta2016theory,devine2016theory,baron1985does}.
This motivates an evaluation setting where an LLM must track and use a partner’s latent mental state during interaction, 
rather than merely select an option in a fixed text. 
Accordingly, we study ToM in human--AI interaction (HAI), where an LLM agent $A$ interacts with a human $H$ through a multi-turn dialogue. Let $D_{1:t}=(u_1,\ldots,u_t)\in\mathcal{D}$ denote the dialogue history up to turn $t$, where each utterance $u_i$ is produced by either $H$ or $A$. Given a task $T\in\mathcal{G}$, the agent first infers a latent mental state
\begin{equation}
z_{t+1} \sim P_{\mathrm{ToM}}(\cdot \mid D_{1:t}, T), \qquad z_{t+1}\in\mathcal{Z},
\end{equation}
and then generates the next response according to
\begin{equation}
u_{t+1}^A \sim \pi_A(\cdot \mid D_{1:t}, T, z_{t+1}).
\end{equation}
Evaluation is scenario-dependent. For each scenario $\Gamma$, we define a scoring schema $\Gamma=(\Phi_\Gamma,\mathrm{Agg}_\Gamma)$, where $\Phi_\Gamma=\{\phi_j\}_{j=1}^m$ is a set of aspect-wise scoring functions,
\begin{equation}
\phi_j:\mathcal{D}\times\mathcal{G}\times\mathcal{Z}\to[0,1],
\end{equation}
and $\mathrm{Agg}_\Gamma:[0,1]^m\to\mathbb{R}$ aggregates the $m$ aspect scores into a single turn-level score. Let $\tau$ be the dialogue length, and let $w_1,\ldots,w_\tau$ be nonnegative temporal weights satisfying $\sum_{t=1}^{\tau} w_t=1$. The scenario-specific performance of policy $\pi_A$ on task $T$ is defined as
\begin{equation}
\scriptsize
\begin{split}
\mathcal{M}_\Gamma(\pi_A,T)
&=\mathbb{E}_{D_{1:\tau}\sim P(\cdot\mid \pi_A,H,T)}
 \\
& \Bigg[
\sum_{t=1}^{\tau} w_t \cdot \mathrm{Agg}_\Gamma\!\big(
\phi_{1:m}(D_{1:t},T,z_{t+1})
\big)
\Bigg].
\end{split}
\end{equation}
Here, $P(\cdot\mid \pi_A,H,T)$ denotes the distribution over dialogue trajectories induced by the agent policy $\pi_A$, the human interlocutor $H$, and the task $T$.
% \subsection{Key Shift: From Static Reasoning to Interactive Collaboration}
The move from static benchmarks to interactive evaluation introduces two essential shifts:

\paragraph{Perspective}  
In static benchmarks, the model acts as a \emph{third-person observer}, reasoning about a fixed narrative world. In interactive HAI settings, the model becomes an \emph{active participant}, required to anticipate, adapt to, and influence the human’s mental state throughout interactions from the first-person perspective.

\paragraph{Metrics}  
While static settings evaluate models solely by \emph{accuracy} over predefined answers,
interactive HAI settings require a richer metric. In our formulation, evaluation follows the general schema $\mathcal{M}_\Gamma$, which can incorporate metrics such as goal completion rate and human satisfaction.  
Ultimately, this paradigm shift reframes the evaluation of ToM from a measure of static reasoning accuracy to a measure of dynamic collaborative effectiveness.

%% file: Method.tex
\section{Methodology}

\subsection{Adapt ToM Methods for HAI Interaction}

Existing methods for enhancing the ToM capabilities of LLMs can be broadly categorized into three approaches: prompt engineering, fine-tuning, and external module integration. 
As our primary goal is to study how well the existing techniques can improve model ToM capability rather than to build new AI systems with multiple modules,
% the impact of foundational ToM enhancement techniques, 
we select methods from the first two categories. Specifically, Foresee and Reflect (FaR) \citep{FaR}, Perspective Taking (PT) \citep{SimToM(PT)}, Supervised Fine-tuning (SFT) \citep{ExploreToM}, and Reinforcement Learning (RL) \citep{ToM-RL} to conduct our experiments. A systematic review and our selection criteria are in Appendix \ref{app:selection}.

% \begin{figure} % r=右侧; 也可用 l
%   % \vspace{-0.6cm}
%   \centering
%   \includegraphics[width=0.8\linewidth]{img/hitom.png}
%   % \vspace{-0.6cm}
%   \caption{Method performance on the HiToM-first benchmark.}
%   \label{fig:hitom}
%   \vspace{-0.4cm}
% \end{figure}

A key challenge is that while our HAI interaction setting requires first-person dialogues, most existing ToM methods are designed for third-person, multiple-choice tasks. We therefore adapt the selected methods to be suitable for direct interaction, as illustrated in Figure \ref{fig:framework}. For prompting methods, we retain their core principles (e.g., reflection and perspective-taking) and reformulate the prompts for a first-person conversational context. For fine-tuning methods, we convert the training data to a first-person perspective by replacing the protagonist's name with ``I''. We then apply these adapted methods to two widely used base models, GPT-4o and Llama-3.1-8B, to create our suite of test models. Note that the GPT-RL model is not included due to fine-tuning limitations. Appendix \ref{app:shift} shows that these techniques improve both base models' performance on existing ToM benchmarks.
Next, model variants with these techniques are applied in our interactive evaluations to verify whether improvements on existing benchmarks translate to tangible benefits in dynamic HAI interaction.
% To validate that our adaptations do not compromise the methods' core effectiveness, we first evaluate them on the HiToM-first benchmark, which is a variant of HiToM, applying the perspective shifting method used in fine-tuning. As Figure \ref{fig:hitom} shows, the adapted models perform effectively on this story-based task. More experiments (Appendix \ref{app:shift}) show the same pattern. 
% This observation confirms that the models' ToM reasoning is improved measured by existing benchmarks, 
% Then,  our exploration for the primary research question: \textit{can demonstrated ToM improvements translate to tangible benefits in dynamic HAI interaction?}

\subsection{Identify HAI Interaction Scenarios}
% Our study aims to cover different HAI interaction scenarios for comprehensiveness.
Before experiments, we identify HAI interaction scenario types to guide what datasets and metrics should be used for a comprehensive evaluation.
Interaction Process Analysis (IPA) shows that human group interaction reliably bifurcates into task and socio-emotional processes \cite{bales1950interaction}. 
Driven by this classic theory, we classify the HAI scenarios into two distinct categories: goal-oriented and experience-oriented. 
\paragraph{Goal-oriented tasks}  
Tasks in this categoty involve users leveraging an LLM as an \textit{assistant} to accomplish a specific and measurable objective (e.g., code generation and document editing). \revise{Prior research suggests that ToM can improve task accuracy by strengthening the coordination protocol between the user and the model~\cite{goal_2}. In particular, stronger mental state attribution helps the model infer the user's latent intentions behind underspecified prompts, thereby reducing misinterpretation and improving collaborative execution. This view is further supported by evidence that collective intelligence depends more strongly on social sensitivity than on the individual IQ of group members, highlighting the importance of mental state attribution for collaborative accuracy~\cite{goal_3}. Related work also identifies ToM as a foundational mechanism that enables AI systems to move beyond rigid tool-like behavior and act instead as adaptive partners in dynamic and ambiguous interactions~\cite{goal_1}. The effectiveness of such ToM-driven collaboration is ultimately reflected in objective external outcomes, such as accuracy, pass@k, and overall task success.}
% Within this framework, ToM becomes integral to effective coordination, as success hinges on correctly attributing the partner's intentions and knowledge state. The efficacy of this ToM-driven collaboration is reflected in objective, external outcomes such as accuracy, pass@k, and task success. This aligns with findings that ToM-linked social sensitivity predicts group problem-solving success in both face-to-face and online environments \citep{woolley2010collective,engel2014rme}.
\paragraph{Experience-oriented tasks} 
Tasks under this class aim to cultivate a high-quality relational experience, including gaining emotional support, engaging in creative exploration, or achieving intellectual satisfaction. 
\revise{Previous studies suggest that incorporating ToM can substantially improve such interactions by enabling LLMs to move beyond surface-level semantic matching and reason more explicitly about users' underlying mental and emotional states. In particular, ToM-aware agents have been shown to better capture the latent beliefs, desires, and intentions that drive a counterpart's behavior, leading to more socially grounded and empathetic responses rather than merely reactive ones~\cite{experience1}. Related work further shows that explicitly infusing ToM into socially intelligent LLM agents improves dialogue effectiveness not only in immediate exchanges, but also in long-horizon interactions that require strategic adaptation and relationship maintenance over time~\cite{experience2}. Accordingly, the value of ToM in experience-oriented settings is reflected primarily in qualitative interaction outcomes, such as users' sense of being understood, perceived partnership, relational quality, and overall engagement.}

\subsection{Evaluate ToM in HAI}
We aggregate four real-world datasets to facilitate comprehensive ToM evaluation. Specifically, to assess performance on \textit{goal-oriented tasks}, we select two benchmarks that simulate real-world collaborative problem-solving: 1) \textit{ChatBench} \citep{Chatbench}, which reframes the MMLU dataset into conversational interactions covering subjects like math, physics, and moral reasoning. Performance is measured by the accuracy of the final answer derived from the human-AI interaction. 2) \textit{CollabLLM} \citep{Collabllm}, which studies multi-turn human-LLM collaboration. We adopt its evaluation pipeline for code generation (BigCodeBench) and document editing (MediumDocEdit), using pass rate and BLEU scores as the respective metrics. 

In the realm of \textit{experience-oriented tasks}, our evaluation centers on two datasets designed to assess an LLM's ability to provide empathetic support. 1)  \textit{MentalChat16K} \citep{Mentalchat16k} offers a rich collection of conversations in a mental health counseling context, covering conditions like depression and anxiety. 2) \textit{Emotional-Support-Conversation} (ESC) \citep{ESC} focuses more broadly on emotional support scenarios. Due to their thematic overlap, we apply a unified set of evaluation metrics (e.g., open-mindness and emphathy) to both datasets following MentalChat16K.

% Our benchmark investigates the effectiveness of ToM enhancement techniques across nine diverse tasks. Beyond these automated metrics, we also conduct a crowdsourcing user study to verify our findings. By combining objective scores with human feedback, we ensure our results accurately capture genuine user perceptions.
% \paragraph{Setup} 
Beyond the simulated benchmarking, we also conduct a crowdsourcing user study to verify our findings. \revise{We only have human evaluation on experience-oriented tasks because these tasks depend on subjective user experience, making human judgment essential, whereas goal-oriented tasks can already be reliably assessed through established simulations.}
% Our benchmarking and case studies reveal that although these ToM enhancement methods fail on goal-oriented tasks, they demonstrate notable potential for enhancing empathetic communication. 
% Our benchmarking and case studies reveal that ToM enhancement methods demonstrate notable potential for enhancing empathetic communication. 
% To further validate it and deeply understand its impact on real user experience, we recruited 100 participants from Prolific \citep{Prolific} to evaluate ToM methods on six experience-oriented tasks.
% \haotian{We need to discuss why not considering goal-oriented tasks.}
\begin{table*}[t]

\centering
\small
\resizebox{0.95\textwidth}{!}{
\begin{tabular}{lllllll}
\toprule
\textbf{Model} & \textbf{Elem Math} & \textbf{HS Math} & \textbf{College Math} & \textbf{Moral} & \textbf{Physics} & \textbf{Overall} \\
\midrule
\rowcolor{blue!10} Llama-3.1-8B
  & 85.16 
  & \textbf{64.59}
  & 44.47  
  & 72.26 
  & 74.76  
  & \textbf{71.38}  \\
\midrule
Llama-3.1-8B-FaR
  & \textbf{86.53} \textcolor{green!60!black}{(+1.37)} 
  & 64.32 \textcolor{red!70!black}{(-0.27)} 
  & 46.84 \textcolor{green!60!black}{(+2.37)} 
  & 67.62 \textcolor{red!70!black}{(-4.64)} 
  & 75.24 \textcolor{green!60!black}{(+0.48)} 
  & 70.98 \textcolor{red!70!black}{(-0.40)} \\
Llama-3.1-8B-PT
  & 83.79 \textcolor{red!70!black}{(-1.37)} 
  & 63.37 \textcolor{red!70!black}{(-1.22)} 
  & 43.16 \textcolor{red!70!black}{(-1.31)} 
  & 69.29 \textcolor{red!70!black}{(-2.98)} 
  & 74.05 \textcolor{red!70!black}{(-0.71)} 
  & 69.85 \textcolor{red!70!black}{(-1.53)} \\
Llama-3.1-8B-SFT
  & 83.05 \textcolor{red!70!black}{(-2.11)} 
  & 62.63 \textcolor{red!70!black}{(-1.96)} 
  & \textbf{48.16} \textcolor{green!60!black}{(+3.69)} 
  & \textbf{73.45} \textcolor{green!60!black}{(+1.19)} 
  & 68.21$^{\dagger}$ \textcolor{red!70!black}{(-6.55)} 
  & 69.62 \textcolor{red!70!black}{(-1.76)} \\
Llama-3.1-8B-RL
  & 85.79 \textcolor{green!60!black}{(+0.63)} 
  & 61.47 \textcolor{red!70!black}{(-3.12)} 
  & 43.42 \textcolor{red!70!black}{(-1.05)} 
  & 71.19 \textcolor{red!70!black}{(-1.07)} 
  & \textbf{76.43} \textcolor{green!60!black}{(+1.67)} 
  & 70.81 \textcolor{red!70!black}{(-0.57)} \\
\midrule
\addlinespace
\rowcolor{green!10} GPT-4o
  & 93.16
  & \textbf{80.32} 
  & 69.21 
  & 76.19  
  & \textbf{88.45} 
  & 83.18  \\
\midrule
GPT-4o-FaR
  & 91.58 \textcolor{red!70!black}{(-1.58)} 
  & 79.89 \textcolor{red!70!black}{(-0.43)} 
  & 69.47 \textcolor{green!60!black}{(+0.26)} 
  & \textbf{80.48} \textcolor{green!60!black}{(+4.29)} 
  & 87.50 \textcolor{red!70!black}{(-0.95)} 
  & \textbf{83.43} \textcolor{green!60!black}{(+0.25)} \\
GPT-4o-PT
  & 92.00 \textcolor{red!70!black}{(-1.16)} 
  & 78.53 \textcolor{red!70!black}{(-1.79)} 
  & 67.11 \textcolor{red!70!black}{(-2.10)} 
  & 78.81 \textcolor{green!60!black}{(+2.62)} 
  & 87.50 \textcolor{red!70!black}{(-0.95)} 
  & 82.63 \textcolor{red!70!black}{(-0.55)} \\
GPT-4o-SFT
  & \textbf{93.58} \textcolor{green!60!black}{(+0.42)} 
  & 79.05 \textcolor{red!70!black}{(-1.27)} 
  & \textbf{70.53} \textcolor{green!60!black}{(+1.32)} 
  & 78.93 \textcolor{green!60!black}{(+2.74)} 
  & 86.67 \textcolor{red!70!black}{(-1.78)} 
  & 83.31 \textcolor{green!60!black}{(+0.13)} \\
\bottomrule
\end{tabular}}
\caption{Performance of  model variations on the ChatBench benchmark,  where $^{\dagger}$ indicates a statistically significant decrease compared with the corresponding base model at $p<0.05$.}
\vspace{-0.4cm}
\label{tab:chatbench}
\end{table*}
In the study, we recruited 100 participants from Prolific \citep{Prolific} to evaluate ToM methods on six experience-oriented tasks (e.g., job crisis, academic pressure).
Participants are randomly assigned to compare variants with different ToM enhancement techniques within either the GPT-4o family or the Llama-3.1-8B family. Each participant chooses a personally resonant task and engages in a three-round conversation. In each round, they rank anonymized and randomized model responses, providing a justification for their choice. The top-ranked response is used to continue the dialogue. 
After three rounds, they provide final qualitative feedback on the overall experience. 
% \paragraph{Setup} 
Details are in Appendix \ref{app:user_ex}. 

% Our benchmark investigates the effectiveness of ToM enhancement techniques across seven diverse tasks. Beyond these automated metrics, we also conduct a crowdsourcing user study to verify our findings. By combining objective scores with human feedback, we ensure our results accurately capture genuine user perceptions.

%% file: Experiment.tex
\section{Results and Findings}

% \FloatBarrier

% \vspace{-1em}
% This section introduces insights into ToM enhancement methods from our interactive evaluations.
% insights into ToM enhancement methods in HAI interaction 
% based on our evaluation framework instantiation with four benchmarks and a real-world user study.
% \vspace{-1em}

% , we introduce the evaluation setup and results of existing ToM enhancement techniques with our evaluation framework in HAI interaction scenarios.
% Our evaluation include synthesized benchmarks and 
\subsection{Goal-Oriented Tasks}
\revise{Based on our statistical analysis results (details in Appendix \ref{app:test}), all methods fail to yield statistically significant improvements.}
% We discuss more details about the results in this section.
% To assess performance on goal-oriented tasks, we select two benchmarks that simulate real-world collaborative problem-solving: 1) \textit{ChatBench} \citep{Chatbench}, which reframes the MMLU dataset into conversational interactions covering subjects like math, physics, and moral reasoning. Performance is measured by the accuracy of the final answer derived from the human-AI interaction. 2) \textit{CollabLLM} \citep{Collabllm}, which studies multi-turn human-LLM collaboration. We adopt its evaluation pipeline for code generation (BigCodeBench-Chat) and document editing (MediumDocEdit-Chat), using pass rate and BLEU scores as the respective metrics.

\paragraph{ChatBench}

As shown in Table \ref{tab:chatbench}, our results indicate that none of the ToM enhancement methods offer a reliable path to improving model performance \revise{with statistical significance} . This limited effectiveness is evident in the overall scores: only GPT-4o-FaR and GPT-4o-SFT achieve marginal gains of up to 0.25, while variants like Llama-3.1-8B-PT and Llama-3.1-8B-SFT experience a significant performance decline of up to 1.76 points. The unpredictable nature of these methods is further highlighted by their volatile performance across different subjects. For example, while Llama-3.1-8B-SFT improves College Math by 3.69 points, its performance on Physics decreases massively by 6.55 points, leading to a failure to enhance the base model overall.
The Moral category, however, appears to be a domain with potential for targeted enhancement, which is particularly relevant to ToM. While three of the GPT-4o variants see significant boosts in this area, the methods fail to produce any effective improvement for the Llama-3.1-8B variants. This discrepancy shows the low situational efficacy of these techniques, as they cannot guarantee positive results on even one targeted domain.
% \haotian{Nanxu, please double check this part, I change a few words.}

\paragraph{CollabLLM}

\begin{figure}[htb]
    \centering
    % \vspace{-0.6cm}
    % Use resizebox to scale the image
      \includegraphics[width=\linewidth]{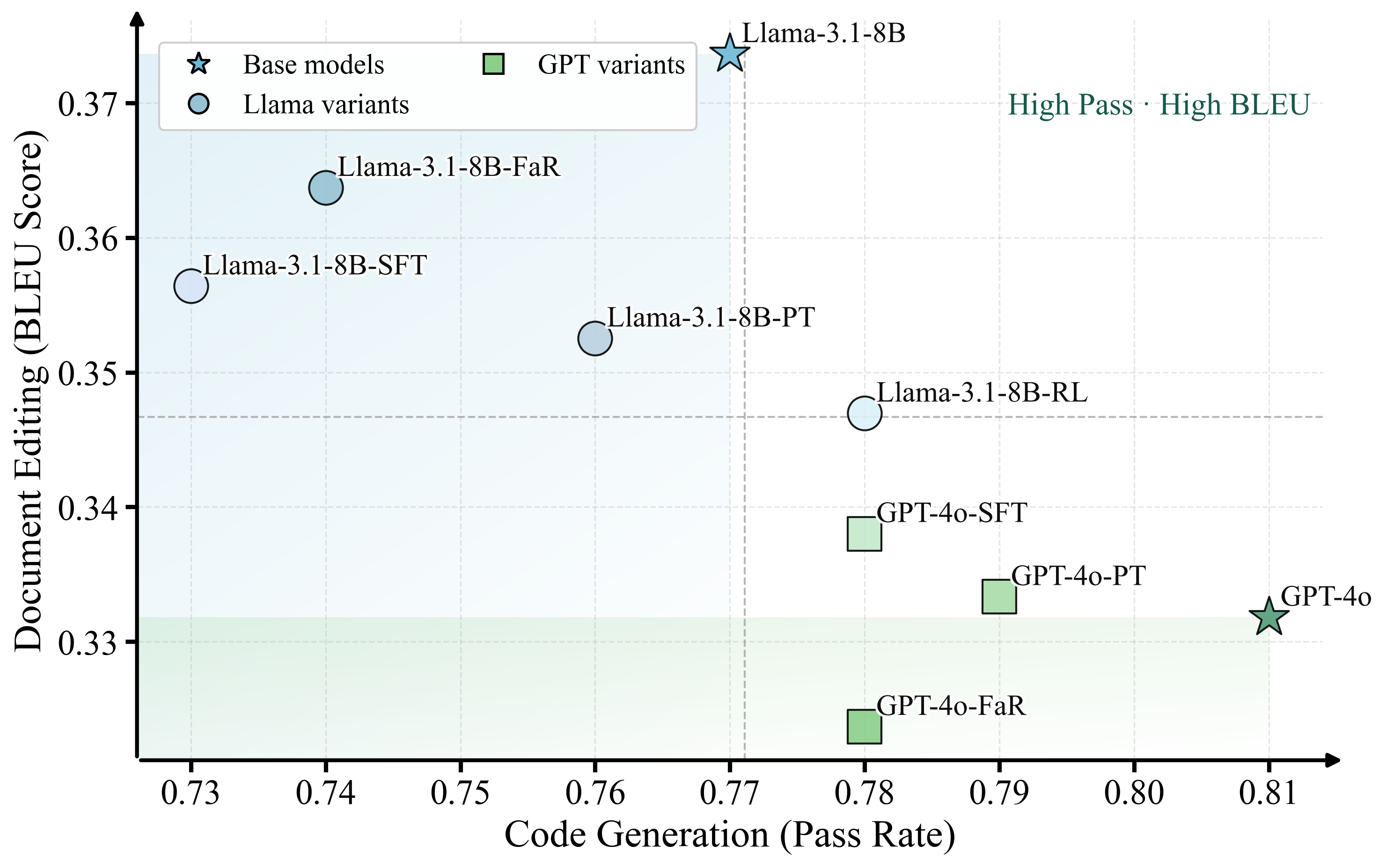}%
    \caption{Model variants' performance on CollabLLM.}
    \label{fig:collabllm}
    \vspace{-0.4cm}
\end{figure}

Figure \ref{fig:collabllm} presents the evaluation results on CollabLLM across document editing and code generation. Generally, these outcomes align with the findings from ChatBench, indicating that the fine-tuning methods do not yield consistent and \revise{statistically significant} improvements on these goal-oriented tasks.
Focusing on document editing, the Llama-3.1-8B baseline acts as the peak performer for its family, with all of its variants failing to match its score. This trend is particularly pronounced for Llama-3.1-8B-RL, which exhibits a performance drop of approximately 0.027. The GPT-4o family shows a slightly more positive, albeit mixed, response; while the SFT and PT variants yield minor benefits, the FaR variant slightly underperforms its baseline.
In the code generation domain, performance degradation is the dominant trend for nearly all variants across both families. The sole exception is Llama-3.1-8B-RL, which achieves a marginal improvement of 0.01. This contrasts sharply with models like Llama-3.1-8B-SFT, which shows a significant performance decrease of approximately 0.04 compared to its baseline.
The findings across two model families jointly highlight the existing ToM enhancement methods' volatile impact.

% \revise{Significance test results indicate that all the methods fail to yield statistically significant improvements. In the CollabLLM and Codebench tests, no variations significantly outperformed the baseline models. Detailed statistical test results can be found in Appendix \ref{app:test}.}

% \begin{figure}[h]
%     \centering
%     \begin{subfigure}[t]{0.47\linewidth}
%         \centering
%         \includegraphics[width=1.65in]{img/case1.pdf}
%         \caption{Goal-Oriented}
%         \label{exp:llm}
%     \end{subfigure}
%     \hfill
%     \begin{subfigure}[t]{0.47\linewidth}
%         \centering
%         \includegraphics[width=1.65in]{img/case2.pdf}
%         \caption{Experience-Oriented}
%         \label{exp:optimizer}
%     \end{subfigure}
%     \caption{Comparison between Goal-Oriented and Experience-Oriented tasks.}
% \end{figure}
% \includegraphics[width=1em]{img/light-bulb.png}\textbf{Takeaway 1: } 

\begin{lem}{}{}%
\vspace{-0.2cm}
% \small
\includegraphics[width=1em]{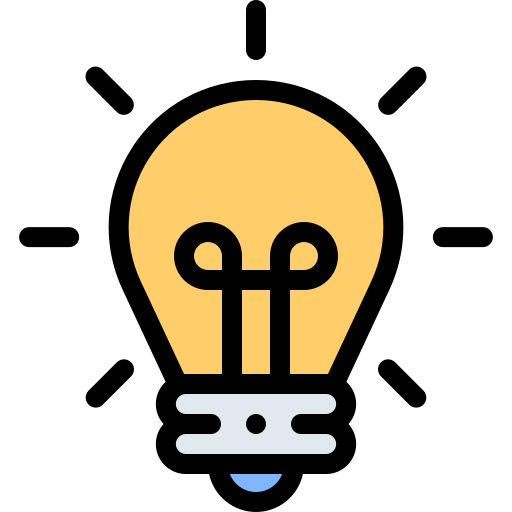}~\textbf{Takeaway 1:} ToM enhancement techniques fail to consistently improve goal-oriented task performance across diverse domains.
% The performance gap between benchmarking methods demonstrates the necessity of introducing a new ToM evaluation paradigm in HAI.
\vspace{-0.1cm}
\end{lem}

\subsection{Experience-Oriented Tasks}

\revise{Different from goal-oriented tasks, the tested methods provide statistically significant improvements for GPT models across all dimensions. On the other hand, fine-tuning with RL or SFT significantly downgrades the performance of the Llama model in Safety and Ethical dimensions.}

% Significance test results indicate that in the realm of emotional and mental health, the tested methods significantly improves the GPT model, but it negatively impacts Llama, which shows a significant decline in Safety and Ethical after being fine-tuned with RL or SFT.}

% In the realm of experience-oriented tasks, our evaluation centers on two key datasets designed to assess an LLM's ability to provide empathetic support. 1)  \textit{MentalChat16K} \citep{Mentalchat16k} offers a rich collection of conversations in a mental health counseling context, covering conditions like depression and anxiety. 2) \textit{Emotional-Support-Conversation} (ESC) \citep{ESC} focuses more broadly on emotional support scenarios. Due to their thematic overlap, we apply a unified set of evaluation metrics to both datasets following MentalChat16K.
% The metrics were synthesized from previous research in evaluating qualities of mental support.
% These metrics include  Active Listening, Empathy \& Validation, Safety \& Trustworthiness, Open-mindedness \& Non-judgment, Clarity \& Encouragement, Boundaries \& Ethical, and Holistic Approach.

\paragraph{MentalChat16K}

\begin{table*}[ht]
\centering
\small
\setlength{\tabcolsep}{4pt}
\resizebox{0.95\textwidth}{!}{
\begin{tabular}{lllllllll}
\toprule
\textbf{Model} &
\textbf{Listening} &
\textbf{Empathy} &
\textbf{Safety} &
\textbf{Open-mind} &
\textbf{Clarity} &
\textbf{Ethical} &
\textbf{Holistic} &
\textbf{Overall} \\
\midrule
\multicolumn{9}{c}{\textit{MentalChat16K}}\\
\midrule
\rowcolor{red!10} Llama-3.1-8B
  & 7.15 
  & 7.04 
  & 7.99 
  & 8.36 
  & 7.54 
  & \textbf{5.85} 
  & 7.67 
  & 7.37  \\
\midrule
Llama-3.1-8B-FaR
  & 7.10 \textcolor{red!70!black}{(-0.05)}
  & 7.14$^{*}$ \textcolor{green!60!black}{(+0.10)}
  & 8.01 \textcolor{green!60!black}{(+0.02)}
  & 8.45$^{*}$ \textcolor{green!60!black}{(+0.09)}
  & 7.67$^{*}$ \textcolor{green!60!black}{(+0.13)}
  & 5.72 \textcolor{red!70!black}{(-0.13)}
  & 7.66 \textcolor{red!70!black}{(-0.01)}
  & 7.39 \textcolor{green!60!black}{(+0.02)} \\
Llama-3.1-8B-PT
  & 7.27$^{**}$ \textcolor{green!60!black}{(+0.12)}
  & \textbf{7.24}$^{***}$ \textcolor{green!60!black}{(+0.20)}
  & \textbf{8.19}$^{***}$ \textcolor{green!60!black}{(+0.20)}
  & \textbf{8.49}$^{**}$ \textcolor{green!60!black}{(+0.13)}
  & \textbf{7.75}$^{**}$ \textcolor{green!60!black}{(+0.21)}
  & 5.80 \textcolor{red!70!black}{(-0.05)}
  & \textbf{7.71} \textcolor{green!60!black}{(+0.04)}
  & \textbf{7.49}$^{***}$ \textcolor{green!60!black}{(+0.12)} \\
Llama-3.1-8B-SFT
  & 7.25$^{*}$ \textcolor{green!60!black}{(+0.10)}
  & 7.05 \textcolor{green!60!black}{(+0.01)}
  & 8.15$^{**}$ \textcolor{green!60!black}{(+0.16)}
  & 8.36 \textcolor{gray}{(0.00)}
  & 7.64 \textcolor{green!60!black}{(+0.10)}
  & 5.58$^{\dagger\dagger\dagger}$ \textcolor{red!70!black}{(-0.27)}
  & 7.48$^{\dagger\dagger\dagger}$ \textcolor{red!70!black}{(-0.19)}
  & 7.36 \textcolor{red!70!black}{(-0.01)} \\
Llama-3.1-8B-RL
  & \textbf{7.33}$^{***}$ \textcolor{green!60!black}{(+0.18)}
  & 7.14$^{*}$ \textcolor{green!60!black}{(+0.10)}
  & 8.07 \textcolor{green!60!black}{(+0.08)}
  & 8.38 \textcolor{green!60!black}{(+0.02)}
  & 7.72$^{**}$ \textcolor{green!60!black}{(+0.18)}
  & 5.50$^{\dagger\dagger\dagger}$ \textcolor{red!70!black}{(-0.35)}
  & 7.54$^{\dagger\dagger}$ \textcolor{red!70!black}{(-0.13)}
  & 7.38 \textcolor{green!60!black}{(+0.01)} \\

\midrule
\addlinespace

\rowcolor{yellow!10} GPT-4o
  & 6.77 
  & 6.52 
  & 8.40 
  & 8.40 
  & 7.54 
  & 6.24 
  & 7.73 
  & 7.37  \\
\midrule
GPT-4o-FaR
  & 7.12$^{***}$ \textcolor{green!60!black}{(+0.35)}
  & 6.85$^{***}$ \textcolor{green!60!black}{(+0.33)}
  & \textbf{8.52}$^{*}$ \textcolor{green!60!black}{(+0.12)}
  & 8.53$^{**}$ \textcolor{green!60!black}{(+0.13)}
  & 7.66$^{*}$ \textcolor{green!60!black}{(+0.12)}
  & 6.42$^{***}$ \textcolor{green!60!black}{(+0.18)}
  & \textbf{7.97}$^{***}$ \textcolor{green!60!black}{(+0.24)}
  & \textbf{7.58}$^{***}$ \textcolor{green!60!black}{(+0.21)} \\
GPT-4o-PT
  & \textbf{7.26}$^{***}$ \textcolor{green!60!black}{(+0.49)}
  & \textbf{6.91}$^{***}$ \textcolor{green!60!black}{(+0.39)}
  & 8.45 \textcolor{green!60!black}{(+0.05)}
  & \textbf{8.54}$^{**}$ \textcolor{green!60!black}{(+0.14)}
  & \textbf{7.70}$^{**}$ \textcolor{green!60!black}{(+0.16)}
  & 6.28 \textcolor{green!60!black}{(+0.04)}
  & 7.89$^{***}$ \textcolor{green!60!black}{(+0.16)}
  & \textbf{7.58}$^{***}$ \textcolor{green!60!black}{(+0.21)} \\
GPT-4o-SFT
  & 6.80 \textcolor{green!60!black}{(+0.03)}
  & 6.56 \textcolor{green!60!black}{(+0.04)}
  & 8.42 \textcolor{green!60!black}{(+0.02)}
  & 8.39 \textcolor{red!70!black}{(-0.01)}
  & 7.45 \textcolor{red!70!black}{(-0.09)}
  & \textbf{6.47}$^{***}$ \textcolor{green!60!black}{(+0.23)}
  & 7.74$^{*}$ \textcolor{green!60!black}{(+0.01)}
  & 7.40 \textcolor{green!60!black}{(+0.03)} \\
\midrule
\multicolumn{9}{c}{\textit{Emotional-Support-Conversation}}\\
\midrule
\rowcolor{red!10} Llama-3.1-8B
  & 7.31 
  & 7.29 
  & 8.09 
  & 8.29 
  & 7.73 
  & 5.92 
  & 7.52 
  & 7.45  \\
\midrule
Llama-3.1-8B-FaR
  & 7.38 \textcolor{green!60!black}{(+0.07)}
  & 7.35 \textcolor{green!60!black}{(+0.06)}
  & 8.02 \textcolor{red!70!black}{(-0.07)}
  & 8.34 \textcolor{green!60!black}{(+0.05)}
  & \textbf{7.75} \textcolor{green!60!black}{(+0.02)}
  & 6.06$^{*}$ \textcolor{green!60!black}{(+0.14)}
  & \textbf{7.63}$^{**}$ \textcolor{green!60!black}{(+0.11)}
  & 7.50 \textcolor{green!60!black}{(+0.05)} \\
Llama-3.1-8B-PT
  & 7.34 \textcolor{green!60!black}{(+0.03)}
  & \textbf{7.45}$^{**}$ \textcolor{green!60!black}{(+0.16)}
  & \textbf{8.14} \textcolor{green!60!black}{(+0.05)}
  & \textbf{8.38}$^{*}$ \textcolor{green!60!black}{(+0.09)}
  & 7.71 \textcolor{red!70!black}{(-0.01)}
  & \textbf{6.08}$^{**}$ \textcolor{green!60!black}{(+0.16)}
  & 7.59$^{*}$ \textcolor{green!60!black}{(+0.07)}
  & \textbf{7.53}$^{**}$ \textcolor{green!60!black}{(+0.08)} \\
Llama-3.1-8B-SFT
  & 7.34 \textcolor{green!60!black}{(+0.03)}
  & 7.31 \textcolor{green!60!black}{(+0.02)}
  & 7.98$^{\dagger}$ \textcolor{red!70!black}{(-0.11)}
  & 8.23 \textcolor{red!70!black}{(-0.06)}
  & 7.74 \textcolor{green!60!black}{(+0.01)}
  & 5.77$^{\dagger}$ \textcolor{red!70!black}{(-0.16)}
  & 7.35$^{\dagger\dagger\dagger}$ \textcolor{red!70!black}{(-0.17)}
  & 7.39$^{\dagger\dagger}$ \textcolor{red!70!black}{(-0.06)} \\
Llama-3.1-8B-RL
  & \textbf{7.40}$^{*}$ \textcolor{green!60!black}{(+0.09)}
  & 7.38$^{*}$ \textcolor{green!60!black}{(+0.09)}
  & 7.97$^{\dagger\dagger}$ \textcolor{red!70!black}{(-0.12)}
  & 8.34 \textcolor{green!60!black}{(+0.05)}
  & 7.70 \textcolor{red!70!black}{(-0.03)}
  & 5.52$^{\dagger\dagger\dagger}$ \textcolor{red!70!black}{(-0.40)}
  & 7.39$^{\dagger\dagger}$ \textcolor{red!70!black}{(-0.13)}
  & 7.39$^{\dagger\dagger}$ \textcolor{red!70!black}{(-0.06)} \\
\midrule
\addlinespace
\rowcolor{yellow!10} GPT-4o
  & 6.92 
  & 6.73 
  & 8.33 
  & 8.27 
  & 7.53 
  & 6.11 
  & 7.64 
  & 7.36  \\
\midrule
GPT-4o-FaR
  & \textbf{7.12}$^{***}$ \textcolor{green!60!black}{(+0.20)}
  & \textbf{6.92}$^{***}$ \textcolor{green!60!black}{(+0.19)}
  & \textbf{8.42}$^{*}$ \textcolor{green!60!black}{(+0.09)}
  & 8.42$^{***}$ \textcolor{green!60!black}{(+0.15)}
  & \textbf{7.72}$^{***}$ \textcolor{green!60!black}{(+0.19)}
  & \textbf{6.32}$^{***}$ \textcolor{green!60!black}{(+0.21)}
  & \textbf{7.86}$^{***}$ \textcolor{green!60!black}{(+0.22)}
  & \textbf{7.54}$^{***}$ \textcolor{green!60!black}{(+0.18)} \\
GPT-4o-PT
  & 7.02$^{*}$ \textcolor{green!60!black}{(+0.10)}
  & 6.89$^{***}$ \textcolor{green!60!black}{(+0.16)}
  & 8.35 \textcolor{green!60!black}{(+0.02)}
  & 8.29 \textcolor{green!60!black}{(+0.02)}
  & 7.55 \textcolor{green!60!black}{(+0.02)}
  & 6.20 \textcolor{green!60!black}{(+0.09)}
  & 7.61 \textcolor{red!70!black}{(-0.03)}
  & 7.42$^{*}$ \textcolor{green!60!black}{(+0.06)} \\
GPT-4o-SFT
  & 7.06$^{**}$ \textcolor{green!60!black}{(+0.14)}
  & 6.87$^{***}$ \textcolor{green!60!black}{(+0.14)}
  & \textbf{8.42}$^{*}$ \textcolor{green!60!black}{(+0.09)}
  & \textbf{8.43}$^{**}$ \textcolor{green!60!black}{(+0.16)}
  & 7.53 \textcolor{gray}{(0.00)}
  & 6.25$^{*}$ \textcolor{green!60!black}{(+0.14)}
  & 7.75$^{**}$ \textcolor{green!60!black}{(+0.11)}
  & 7.47$^{***}$ \textcolor{green!60!black}{(+0.11)} \\
\bottomrule
\end{tabular}}
\caption{Performance of model variations on MentalChat16K and Emotional-Support-Conversation, where $^{*}$, $^{**}$, and $^{***}$ indicate a statistically significant increase compared with the corresponding base model at $p<0.05$, $p<0.01$, and $p<0.001$, respectively. $^{\dagger}$, $^{\dagger\dagger}$, and $^{\dagger\dagger\dagger}$ indicate a statistically significant decrease at the same thresholds.}
\vspace{-0.2cm}
\label{tab:mentalchat-all}
\end{table*}

\revise{As shown in Table \ref{tab:mentalchat-all}, these methods generally improve empathetic communication skills on the MentalChat16K benchmark, with a top overall gain of 0.21 points. However, the results differ substantially between the Llama and GPT families. For the Llama-3.1-8B family, PT is the most effective variant, achieving the best overall score of 7.49, and its improvements on several dimensions, including Listening, Empathy, Safety, Open-mind, Clarity, and the overall score, are statistically significant. Nevertheless, a critical issue emerges in the form of degradation on certain dimensions. In particular, the Ethical score is consistently reduced across most variants, and this decline is statistically significant for both SFT and RL, with RL showing the largest drop of 0.35 points. Moreover, the decreases on the Holistic dimension are also statistically significant for SFT and RL, suggesting that gains in some local conversational skills do not necessarily translate into more balanced overall support quality.
Conversely, the methods appear more robust on GPT-4o. Both FaR and PT achieve the best overall score of 7.58, and these gains are statistically significant. More importantly, FaR yields statistically significant improvements across most dimensions, including Listening, Empathy, Safety, Open-mind, Clarity, Ethical, and Holistic, indicating a broadly consistent enhancement pattern rather than isolated gains on a few attributes. PT also shows significant improvements on multiple dimensions, though its gains are somewhat less uniform. Overall, these results suggest that the methods are more reliably effective on GPT-4o, whereas on Llama-3.1-8B they are accompanied by clearer trade-offs, especially on Ethical and Holistic aspects.}

\paragraph{Emotional-Support-Conversation (ESC)}
\revise{On the ESC benchmark, the evaluated methods show more mixed effects, particularly for the Llama-3.1-8B family. PT delivers the strongest overall improvement, reaching 7.53, and this gain is statistically significant. However, the SFT and RL variants are clearly detrimental, reducing the overall score to 7.39, with both decreases being statistically significant. More importantly, these degradations are accompanied by statistically significant drops on critical dimensions. For SFT, the model shows significant declines on Safety, Ethical, and Holistic. RL exhibits an even stronger negative pattern, with statistically significant decreases on Safety, Ethical, Holistic, and the overall score; among them, the most severe regression is a 0.40-point drop on Ethical. These results suggest that, for Llama-3.1-8B, some alignment methods can improve selected conversational traits while simultaneously introducing meaningful regressions in safety-related and holistic support quality.
In contrast, the methods are substantially more stable and consistently beneficial on GPT-4o. FaR is the strongest variant, achieving the best overall score of 7.54 with a statistically significant gain, while also producing significant improvements across nearly all fine-grained dimensions. PT and SFT also lead to statistically significant overall improvements, although their gains are less comprehensive than those of FaR. Notably, unlike the Llama family, the GPT-4o variants do not exhibit statistically significant degradations on Safety, Ethical, or Holistic. This pattern indicates that while these methods can enhance empathetic communication, their deployment on weaker base models may introduce statistically meaningful safety and ethical trade-offs, whereas stronger models such as GPT-4o appear more robust.}

% \includegraphics[width=1em]{img/light-bulb.png}\textbf{Takeaway 2: } ToM-enhanced models improve empathetic skills, but pose potential issues in Holistic, Ethics and Safety.

% \vspace{-0.6cm}

\paragraph{Case Study}

\begin{figure*}[h]
    \centering
    % Use resizebox to scale the image
    \resizebox{1\textwidth}{!}{%
      \includegraphics{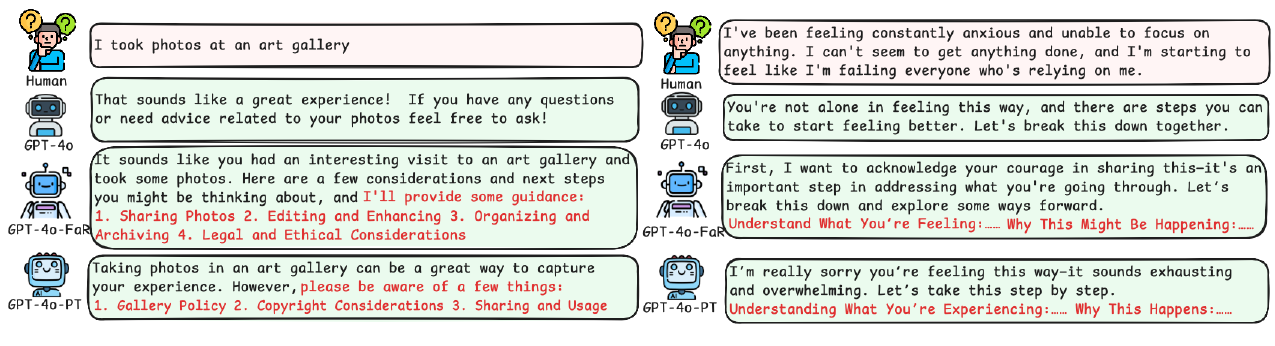}%
    }
    \vspace{-0.6cm}
    \caption{Case studies from a goal-oriented task (left) and a experience-oriented task (right).}
    \label{fig:case}
    \vspace{-0.4cm}
\end{figure*}

To intuitively analyze the behavioral changes that ToM capabilities induce in a model, we present two case studies from ChatBench and MentalChat16K in Figure \ref{fig:case}. Taking the FaR and PT methods as examples, in the case shown on the left, the user makes a simple statement: ``I took photos at an art gallery.'' The base model provides a generic and passive response, such as ``If you have any questions, feel free to ask.'' In contrast, the models with ToM enhancement techniques proactively infer the user's potential intentions, speculating on what the user might implicitly want to ask. This demonstrates that these methods can transform the model's role in a conversation from that of a passive text processor into a proactive listener, who analyzes the underlying users' mental states.

\begin{lem}{}{}%
\vspace{-0.2cm}
% \small
\includegraphics[width=1em]{img/light-bulb.png}~\textbf{Takeaway 2:} ToM enhancements boost empathy and user experience, although fail to support goal achievement. Furthermore, SFT and RL can amplify safety and ethical risks.
\vspace{-0.1cm}
% \vspace{-1em}
\end{lem}

\subsection{User Study}
\label{sec:user}
\begin{table}
% \vspace{-0.5cm}

% \vspace{-0.5cm}
\small
\centering
\resizebox{0.45\textwidth}{!}{
\begin{tabular}{lcccccc}
\toprule
\multirow{2}{*}{\textbf{Method}} & \multicolumn{3}{c}{\textbf{GPT-4o}} & \multicolumn{3}{c}{\textbf{Llama-3.1-8B}} \\
\cmidrule(lr){2-4} \cmidrule(lr){5-7}
 & Mean  & Std & Top-1\% & Mean & Std & Top-1\% \\
\midrule
PT   & 2.43 & 1.09 & 26.5 & 2.88  & 1.42 & 23.2 \\
FaR  & 2.48 & 1.14 & 29.1 & 2.97  & 1.49 & 23.8 \\
SFT  & 2.56 & 1.14 & 22.5 & 2.98  & 1.43 & 22.5 \\
RL   & --   & --   & --   & 3.08  & 1.33 & 11.3 \\
\midrule
\rowcolor{gray!10}
Base & 2.53 & 1.09 & 21.9 & 3.09 & 1.39 & 19.2 \\
\bottomrule
\end{tabular}}
\caption{Overall ranking of ToM methods across GPT and Llama families (lower is better). Top-1 (\%) indicates the proportion of times ranked first. }
\vspace{-0.4cm}
\label{tab:overall_ranking}
\end{table}

% To further examine human perception of the usefulness of ToM enhancement methods, we conducted a crowdsourcing user study. 

% \begin{figure}
%     \centering
%     % \vspace{-0.6cm}
%     % Use resizebox to scale the image
%       \includegraphics[width=0.8\linewidth]{img/word_cloud.png}%
%     \caption{The word cloud for participants' justification of response preferences.}
%     \label{fig:word_cloud}
%     % \vspace{-0.5cm}
% \end{figure}

\begin{figure*}[h]
    \centering
    % Use resizebox to scale the image
    \resizebox{1\textwidth}{!}{%
      \includegraphics{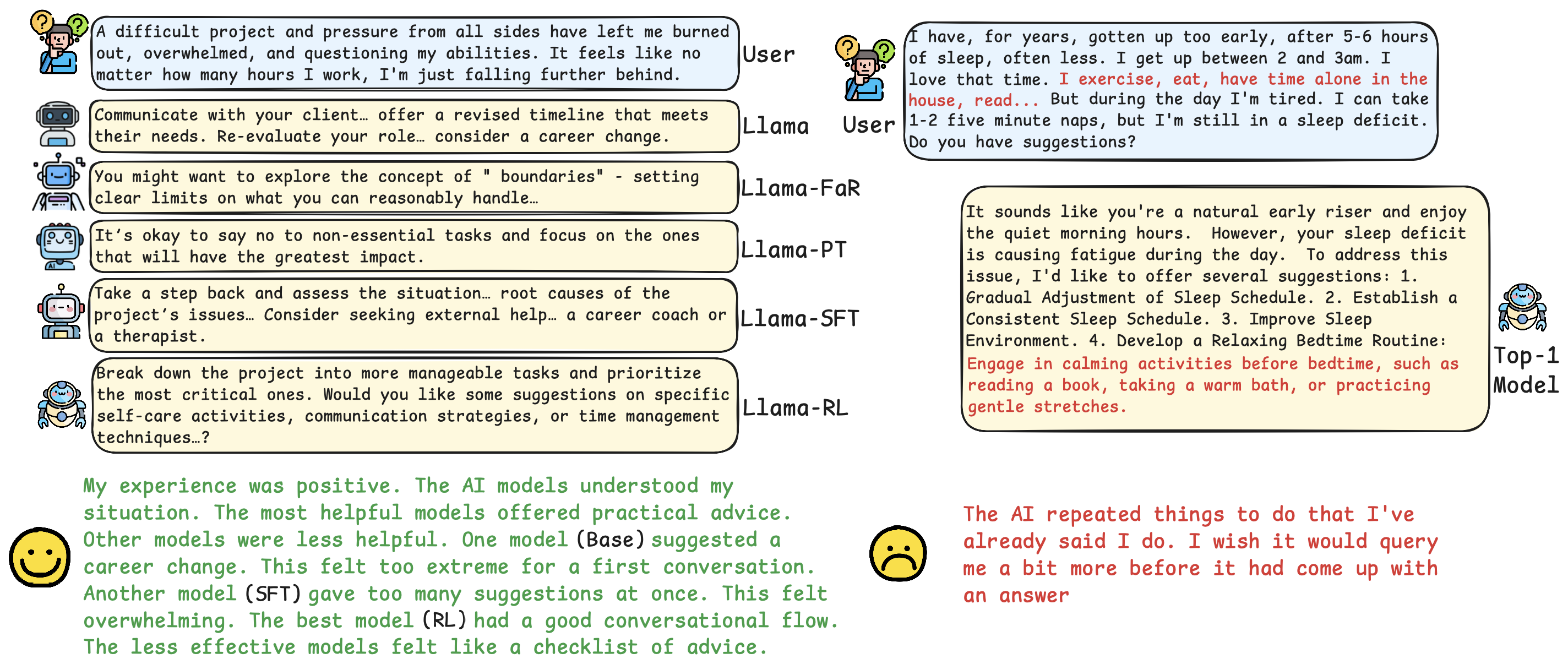}%
    }
    % \vspace{-0.6cm}
    \caption{Cases of positive (left) and negative (right)  experiences and corresponding comments in our user study.}
    % , together with corresponding participants' comments on different methods. }
    \label{fig:user_case}
    % \vspace{cm}
    \vspace{-0.2cm}
\end{figure*}

% \paragraph{Results} 
Our human evaluation reveals a consistent but not \revise{statistically significant preference} for models with ToM enhancement techniques, aligning with the results of experience-oriented benchmarks.
Across two model families, we can see that models based on prompt-based methods (FaR and PT) outperform the base model and the models after fine-tuning, suggesting the potential robustness of prompt-based methods in more diverse real-world user needs. \revise{To further quantify participant-perceived ranking differences, we conduct statistical significance tests on the user study results and compute effect sizes under the Friedman test with Kendall's \(W\) framework. The observed concordance values are extremely small, with \(W=0.0152\) for the GPT group (\(p=0.52\)) and \(W=0.00394\) for the Llama group (\(p=0.94\)). These results indicate negligible agreement among participants, suggesting that the differences between the compared methods are too subtle to be consistently perceived in human evaluation.}
% However, as the results
% To further reveal users' thoughts over methods, we present a word cloud to summarize their expressed opinions (Figure ~\ref{fig:word_cloud}) and two detailed cases (Figure~\ref{fig:case}).
% % More details are available in Appendix~\ref{app:case}.
% In the word cloud,  ``all good'' and ``all helpful'' frequently appear in their comments.
% It partially explains the trivial difference between model variants in our quantitative results: for most of the experience-oriented tasks, the ToM capability of current models might be satisfactory.
The minor differences (such as those described in Figure~\ref{fig:user_case}-left) do not considerably improve experiences in real-world HAI interactions.
% as the participant addressed ``My experience was positive. The AI models understood my situation.''
Another reason for the minor ranking difference lie in diverse conversation goals and personal requirements for LLMs, leading to divergent preferences on models (Figure~\ref{fig:tom_wordcloud} in Appendix~\ref{app:user_results}).
% It is also supported by the case on the right, we can notice that the user is overall satisfied with all responses, possibly implying that all models can infer their problem with their ToM capability and provide appropriate responses.
% The differences between model variants are quite trivial to be sensed by users.

% Though the overall results are optimistic, we must be aware that LLMs' ToM capability is not perfect.
% However, that does not mean that the ToM capability of current models is 
To summarize, we consider that the enhanced models still lack of sufficient ToM capability to capture users' nuanced intention from interactions.
% For example, we noticed that all model variants, including
% the top-1 model identified by the user,
For example, we noticed that all model variants, including
the best model ranked by the user (i.e., Llama-3.1-8B-FaR),
fail in suggesting more diverse methods to facilitate their sleep problem (Figure~\ref{fig:user_case}-right).
Beyond direct instructions for models, the underlying adaption to user preferences and scenarios, such as conversation styles, also poses higher requirements on the ToM capability of LLMs.
Furthermore, our results show that no one model variant can achieve the best across scenarios, implying their limited ToM capability for dynamic and diverse HAI interactions (Table~\ref{tab:task_ranking} in Appendix~\ref{app:user_results}). \revise{Taken together, these results suggest that current ToM-enhancement methods do not yet yield substantial improvements in perceived user experience in realistic human-AI interactions, highlighting the need for more realistic evaluation frameworks and for new approaches to ToM improvement.}

\begin{lem}{\includegraphics[width=1em]{img/light-bulb.png}~\textbf{Takeaway 3}}{}%
\vspace{-0.2cm}
% \small
\includegraphics[width=1em]{img/light-bulb.png}~\textbf{Takeaway 3:} While the ToM methods benefit HAI, realizing their full potential requires enhancing dynamic user understanding.
\vspace{-0.1cm}
\end{lem}
% w \includegraphics[width=1em]{img/light-bulb.png}\textbf{Takeaway 3: } Human evaluation indicates a slight yet consistent preference for ToM-enhanced models
% Despite these trends in average rankings, pairwise Wilcoxon tests failed to find any statistically significant differences, as all comparisons for both model families yielded p-values of .05 or greater. We interpret this as evidence that the human perception of ToM improvements, while present, is subtle and attenuated by the diverse expectations and contexts of individual users. This highlights the challenge in creating universally recognizable improvements in complex social reasoning.

\section{Discussion}
% We move beyond the results and discuss the empirical insights found in our experiments.

% \haotian{We need to be more careful about the results. According to our current results, we have three key observations: 1. HiToM (and original story-reading benchmarks in these method papers) tells us that all methods can enhance performance; 2. Task-oriented benchmarks tell us that all methods cannot increase human-AI task accomplishment performance; 3. Experience-oriented benchmarks tell us that the methods can effectively enhance human user experience, as supported by user study results.  }

% \haotian{To combine them together, we can have three conclusions: 1. (from the angle of the effectiveness of ToM-enhancement methods) Current methods do not show merits in improving collaboration performance but can enhance user experience in counseling; 2. (from the angle of benchmark effectiveness) We can clearly see that the trend of ToM capability in HiToM does not agree with the trend of either goal-oriented or experience-oriented benchmarks or user study results. It reveals the gap between story-reading benchmark methods and our new benchmark methods, suggesting that the real-world effectiveness of ToM enhancement methods cannot be purely represented by story-reading benchmark; 3. (from our evaluation framework perspective) Goal-oriented and experience-oriented tasks are both frequent in HAI. They should be separately evaluated based on user needs, calling for more attention for diverse ToM benchmark datasets. }

\paragraph{HAI symbiosis poses new challenges for ToM.} 
Our evaluation framework marks a methodological shift designed to assess ToM for the challenges of HAI symbiosis.  
% Specifically, we move beyond traditional, static benchmarks that measure a model's third-person analytical ToM, and instead introduce a dynamic, interactive setting that measures its applied, first-person ToM during live conversation. 
This new perspective reveals a significant performance gap.  
We observe that the methods that improve story-reading benchmark performances only show limited and inconsistent benefit in our interactive evaluation.
% methods which yield strong improvements on the story-reading benchmarks, such as Llama-3.1-8B-RL on HiToM-first, do not show a correspondingly clear benefit in our interactive evaluation, where their positive impact is far more subtle. 
This gap highlights the necessity and importance of our framework for gaining a complete picture of a model's true capabilities.  It shows that excelling at test-taking tasks does not guarantee readiness for interactive collaboration.  Therefore, it is essential to complement existing benchmarks with dynamic and interactive evaluations in HAI contexts. 

\paragraph{Enhanced ToM fails to generalize from assistance to companionship.} Our findings show that ToM-enhancement methods improve performance in experience-oriented scenarios but fail in goal-oriented tasks. This performance difference appears to stem from the distinct nature of these two task categories.  Experience-oriented tasks are largely defined by their focus on interpersonal dynamics and responding to affective states like emotions and desires.  In contrast, goal-oriented tasks can require
understanding users' intention progress and underlying knowledge states for task accomplishment.
% achieving concrete objectives, where success is tied to functional outcomes.  
This suggests that ToM proficiency in one type of task may not guarantee success in the other, as each emphasizes different aspects of user understanding.  
This capability gap highlights that future research
% a critical limitation in current evaluation paradigms and ToM enhancement methods.   
% To cultivate a more comprehensive and balanced ToM, the community 
needs diverse and real-world benchmarks that assess a full spectrum of abilities from empathetic support to goal-driven collaboration.
% more diverse benchmarks that assess the full spectrum of abilities, from passive social understanding to active, goal-driven collaboration.

\paragraph{Users require threshold-crossing ToM improvements.} 
Comprehensive ToM capability is fundamental to achieving true HAI symbiosis, as it is the capability that transforms models from passive text processors into proactive, collaborative partners. However, our user study reveals that even these limited gains are not strongly perceived by users, failing to translate into a clear preference. We consider two potential reasons. First, models' ToM capability is satisfactory for a majority of tasks, making the marginal gains often fall below a user's perceptual threshold.
Furthermore, current methods are largely designed for static, story-reading benchmarks and are thus ill-suited for understanding dynamic and nuanced user goals and preferences in live interaction.
% current methods are largely designed for static, story-reading benchmarks and are thus ill-suited for the dynamic demands of live interaction. Second, the marginal gains they produce often fall below a user's perceptual threshold, an effect compounded by the diversity of human preferences. 
Therefore, the path forward requires designing new enhancement methods to understand the nuanced user mental states dynamically in HAI scenarios. Only by optimizing for the complexities of live interaction can we transfer the model improvements from benchmarks to a meaningfully better user experience.

%% file: Related.tex
\section{Related Work}

\paragraph{Assessment of ToM} ToM assessment in LLMs has primarily relied on story-based benchmarks extending classical psychological tests \citep{survey1,survey2}. Early benchmarks like ToMi and Hi-ToM expanded this approach with diverse narratives and higher-order reasoning \citep{ToMi,HiToM}. Subsequent efforts improved protocols, such as ToMChallenges’ varied templates and FANTOM’s detection of “illusory ToM” in dialogues \citep{ToMchallenges, FANToM}. Concurrently, BigToM and OpenToM broadened the scope to include mental states like percepts and emotions \citep{BigToM, OpenToM}. Recent works address domain-specific reasoning (NegotiationToM) and systematic coverage (ToMBench) \citep{NegotiationToM, ToMBench}, alongside novel data generation techniques utilizing search algorithms or information asymmetry \citep{ExploreToM, ToMATO}. However, most benchmarks remain passive evaluations, positioning models as observers. This limitation results in only a partial view of ToM competence, motivating our interactive protocol development.

\paragraph{Enhancement of ToM} Recent research studies enhancing LLM ToM capabilities through three primary categories \citep{survey_assess_enhance}. 1) 
\textit{Prompt engineering} guides reasoning without retraining \citep{MP,TimeToM}. For instance, FaR prompts reflection on predicted story evolutions \citep{FaR}, while SimToM filters context to strictly match a character's perception \citep{SimToM(PT)}.
2) \textit{Fine-tuning} adapts models using specialized datasets. Approaches include ToM-RL, which utilizes reinforcement learning (e.g., GRPO) \citep{ToM-RL}, and ExploreToM, which applies supervised fine-tuning on diverse, challenging benchmarks \citep{ExploreToM}.
3) \textit{External module integration} augments models via specialized components \citep{DWM,VToM,DecomposeToM,Thought-Tracing}. For example, AutoToM refines agent models via inverse planning \citep{AutoToM}. 

%% file: Conclusion.tex
\section{Conclusion}
% In this paper, we reframe ToM evaluation for HAI symbiosis by replacing static, third-person  perspective quizzes with interactive conversations from the first-person perspective. 
% To comprehensively study the effectiveness of the methods for improving ToM, we summarize the HAI scenarios into goal-oriented tasks and experience-oriented tasks. From a systematic benchmarking evaluation and crowdsourcing user study, we find
% that ToM-enhancement methods fail to produce meaningful improvements in goal-oriented collaboration, yet they demonstrate a slight positive impact on experience-oriented interaction. Beyond the experiments, we derive key insights for future evaluation and method design, advocating for a shift towards more practical, context-aware ToM research.
% that acknowledges the different demands of these distinct application scenarios.

% To comprehensively study the effectiveness of the methods for improving ToM, we reframe ToM evaluation for HAI symbiosis by replacing static, third-person perspective quizzes with interactive conversations from the first-person perspective. We propose a novel evaluation framework, evaluating ToM on both goal-oriented and experience-oriented scenarios. From the benchmark and user study results, we reveal three key insights on the future ToM evaluation and improvement: 1) HAI symbiosis poses new challenges for ToM. 2) Enhanced ToM fails to generalize from assistance to companionship. 3) Users require threshold-crossing ToM improvements.
In this paper, we re-examine the effectiveness of ToM enhancement by moving beyond static, third-person evaluation to dynamic, open-ended, first-person human-AI interaction. Built on the HAI symbiosis paradigm, we organize application scenarios into goal-oriented and experience-oriented tasks, and conduct both simulated benchmarking and a user study to assess current ToM enhancement methods in these settings. 
By systematically comparing representative enhancement methods with two distinct base models on nine tasks, our results show that existing evaluation protocols and method designs remain misaligned with real-world human needs and fail to deliver meaningful improvements in user experience. 
% More broadly, these findings are observed across representative LLM-intrinsic enhancement methods, two distinct base models, and diverse task settings, suggesting that the gap between benchmark gains and practical human-AI benefit is not limited to a single method or model family. 
We hope this work provides a guidance for future human-centric ToM evaluation and the development of socially intelligent AI systems for human-AI symbiosis.
% \haotian{Please update the conclusion based on abstract, introduction, and the new Sec. 2-4 organization.}

\section*{Limitations}
Our research is not without limitations. 
First, we only tested a limited coverage of ToM enhancement methods. 
% Though we attempted to cover the most representative methods, some other ones are not considered.
The reasons hindering us from including other methods include (1) it is challenging to ask users to compare many model responses generated by variants with different techniques at the same time and (2) many ToM-enhancement methods are hard to be adapted to HAI scenarios.
For example, TimeToM~\cite{TimeToM} designed an algorithm to model characters' movements and mental states from the third-person perspective, which is challenging to be generalized to open-ended HAI interactions from a first-person perspective.
% character-based stories in ToM evaluation and it is hard to be adapted in HAI scenarios.
% their tailored design for specific evaluation methods.
% For example, 
% Another consideration is that it is challenging to ask users to compare many model responses generated by variants with different techniques at the same time.
As a result, we finally kept the four carefully selected representative techniques that can be adapted to HAI scenarios from both prompt engineering-based and finetuning-based methods.
% \haotian{Nanxu, please fill in with details.}
Second, our research can be extended in more potential HAI scenarios.
Currently, we only gathered data from nine scenarios in four datasets, including math problem solving, collaborative writing, mental counseling, and emotional support.
It is limited by the availability of data from other highly related scenarios, such as customer support.
We sincerely hope to further extend our research when more real-world data is available.
\revise{Finally, rather than compartmentalizing social intelligence into specific metrics, the central aim of our research is to assess the practical utility of methods for improving ToM within human-AI interaction. Traditional datasets tend to evaluate cognitive skills in a standalone, test-taking method. However, genuine interactive environments are highly interwoven, requiring various inferential abilities to operate in concert. It is crucial to obtain an overview of ToM enhancement methods’ proficiency first before delving into diagnosing sub-dimensions. Consequently, this paper deliberately researches whether and how general ToM proficiency improvements benefit human-AI interaction, leaving component-level breakdowns to subsequent research. }

% Our limitations are mainly two-fold. Firstly, because many existing methods are designed specifically for certain benchmarks or use elaborately designed prompts, we cannot apply them in our first-perspective and dynamic scenarios. Secondly, the datasets we used are all open-source.

\section*{Ethics Considerations}
This research involved a user study with 
an IRB-approved study protocol.
The participants were recruited from Prolific. 
All study participants provided informed consent and were compensated for their time. 
They could withdraw from the study at any time when they felt uncomfortable or unwilling to continue.
Their data was anonymized to protect their privacy.
We further checked that there is no personal identifiable information of participants revealed in the paper.

\section*{Acknowledgments}
This research is part of the AFMR collaboration supported by Microsoft Research. 
% All case studies shown in this paper have been checked and processed to ensure no personal identifiable information is revealed. 

%% file: Appendix.tex
\appendix

\section{Method and Result Details}
% \section{Code Availability}
Our code and data are publicly available at \url{https://nanxugong.github.io/ToM-HAI/}. This section introduces our experimental setup and statistical test results for benchmarks.

% Our code and data is publicly available at 
\subsection{ToM Enhancement Method Selection}
\label{app:selection}

\begin{table*}[h]

\centering
\small
\begin{tabular}{p{3.2cm} p{3.7cm} p{3cm} p{2.0cm}}
\toprule
\textbf{Method} & \textbf{Category} & \textbf{Core Idea} & \textbf{Modality} \\
\midrule
Discrete World Models & external module integration & decomposition & Text \\
Metacognitive Prompting & prompt & reflection & Text \\
PercepToM & external module integration & perspective-taking & Text \\
TimeToM & prompt & timeline & Text \\
SimToM & prompt & perspective-taking & Text \\
FaR & prompt & reflection & Text \\
ExploreToM  & finetune & SFT & Text \\
ToM-RL & finetune & RL & Text \\
VToM & external module integration & visual reasoning & Multimodal \\
COKE / COLM & finetune & SFT & Text \\
Thought-Tracing & external module integration & Monte Carlo & Text \\
AutoToM & external module integration & BIP & Multimodal \\
Decompose-ToM & external module integration & decomposition & Text \\
I Cast Detect Thoughts & external module integration & RL dialog & Text \\
\bottomrule
\end{tabular}
\caption{Summary of recent Theory of Mind (ToM) related papers by category, sub-category, and modality.}
\label{tab:tom_papers}
\end{table*}

We firstly review methods for improving the ability of ToM as Table \ref{tab:tom_papers}. 
1) \textit{Discrete World Models (DWM)} \citep{DWM} discretizes narratives into a finite set of belief states and transitions; defines task complexity as the minimal number of states required, and performs stepwise belief updating within this discrete state space.
2) \textit{Metacognitive Prompting (MP)} \citep{MP} embeds a five-phase metacognitive control loop into the prompt—identifying knowns/unknowns, hypothesizing, checking evidence, and revising—so that reasoning is executed as a procedural self-monitoring routine.
3) \textit{PercepToM} \citep{PrecepToM} adopts a two-stage setup: first explicitly annotates each agent’s perceptual availability, then infers beliefs along the perception→belief mapping under that annotation.
4) \textit{TimeToM} \citep{TimeToM} constructs Temporal Belief State Chains (TBSCs) for each character and uses a tool-augmented belief solver to update and query beliefs along an explicit timeline.
5) \textit{SimToM} (Perspective-Taking) \citep{SimToM(PT)} applies two-step prompting: filters the context to the target character’s accessible knowledge, then answers strictly from that restricted viewpoint.
6) \textit{FaR} \citep{FaR} implements a forecast–reflect prompting routine: samples plausible future trajectories of the story, then reflects over these trajectories to select the response or action.
7) \textit{ToM-RL} \citep{ToM-RL} fine-tunes the language model with reinforcement learning (e.g., RLHF/PPO), using ToM-aligned reward signals to optimize generation, optionally preceded by supervised warm-start.
8) \textit{VToM} \citep{VToM} builds a multimodal pipeline that retrieves key video frames, forms a video–text graph, and performs conditional reasoning over this graph to answer belief/intent queries.
9) \textit{COKE} \citep{Coke} constructs a cognitive knowledge graph of structured social/causal chains and conditions or fine-tunes a generator on these chains to enforce cognitively grounded reasoning.
10) \textit{Thought-Tracing} \citep{Thought-Tracing} uses a sequential Monte Carlo–inspired, inference-time procedure that generates, weights, and resamples natural-language hypotheses of agents’ mental states over narrative time.
11) \textit{AutoToM} \citep{AutoToM} leverages automated Bayesian inverse planning: proposes an initial BToM, estimates likelihoods/posteriors via simulation with an LLM-backed proposer, and iteratively refines the model under uncertainty.
12) \textit{I Cast Detect Thoughts} \citep{Decect_Thoughts} trains dialogue policies in a Dungeons-and-Dragons–style interactive environment via RL with ToM-aware rewards, aligning guidance utterances with inferred player intents and world state.
13) \textit{Decompose-ToM} \citep{DecomposeToM} implements a simulation-based task-decomposition pipeline—subject identification, question reframing, world-model update, and knowledge-availability checks—then generates answers from the decomposed reasoning states.

Our method selection follows a two-step protocol. (1) We restrict attention to methods that directly enhance an LLM’s capabilities (via prompting or parameter updates), and therefore exclude external module integration methods. (2) For families of methods sharing a core idea, we choose a single representative to avoid redundancy. Consequently, we evaluate four methods: forsee and reflection (FaR) \citep{FaR}, perspective-taking (PT) \citep{SimToM(PT)}, supervised fine-tuning (SFT) \citep{ExploreToM}, and reinforcement learning (RL) \citep{ToM-RL}.

\subsection{Model Setup}
To comprehensively evaluate various methods for enhancing ToM, we selected two representative LLMs: GPT-4o and Llama-3.1-8B. These base models were chosen to cover a range of model scales and access types (closed- and open-source). For the prompt-based methods, FaR and PT, we utilized the specific prompts shown in Figures \ref{fig:prompt1} and \ref{fig:prompt2}. For SFT, we fine-tuned the base models on the ExploreToM-first dataset, which adapts the original data from a third-person to a first-person perspective \citep{ExploreToM}. Similarly, RL, we followed the established ToM-RL pipeline, using the first-person transformed data \citep{ToM-RL}.

\subsection{ToM Enhancement Method Implementation}
\label{app:shift}
\begin{figure*}[h!]
    \centering
    % Use resizebox to scale the image
    \resizebox{0.5\textwidth}{!}{%
      \includegraphics{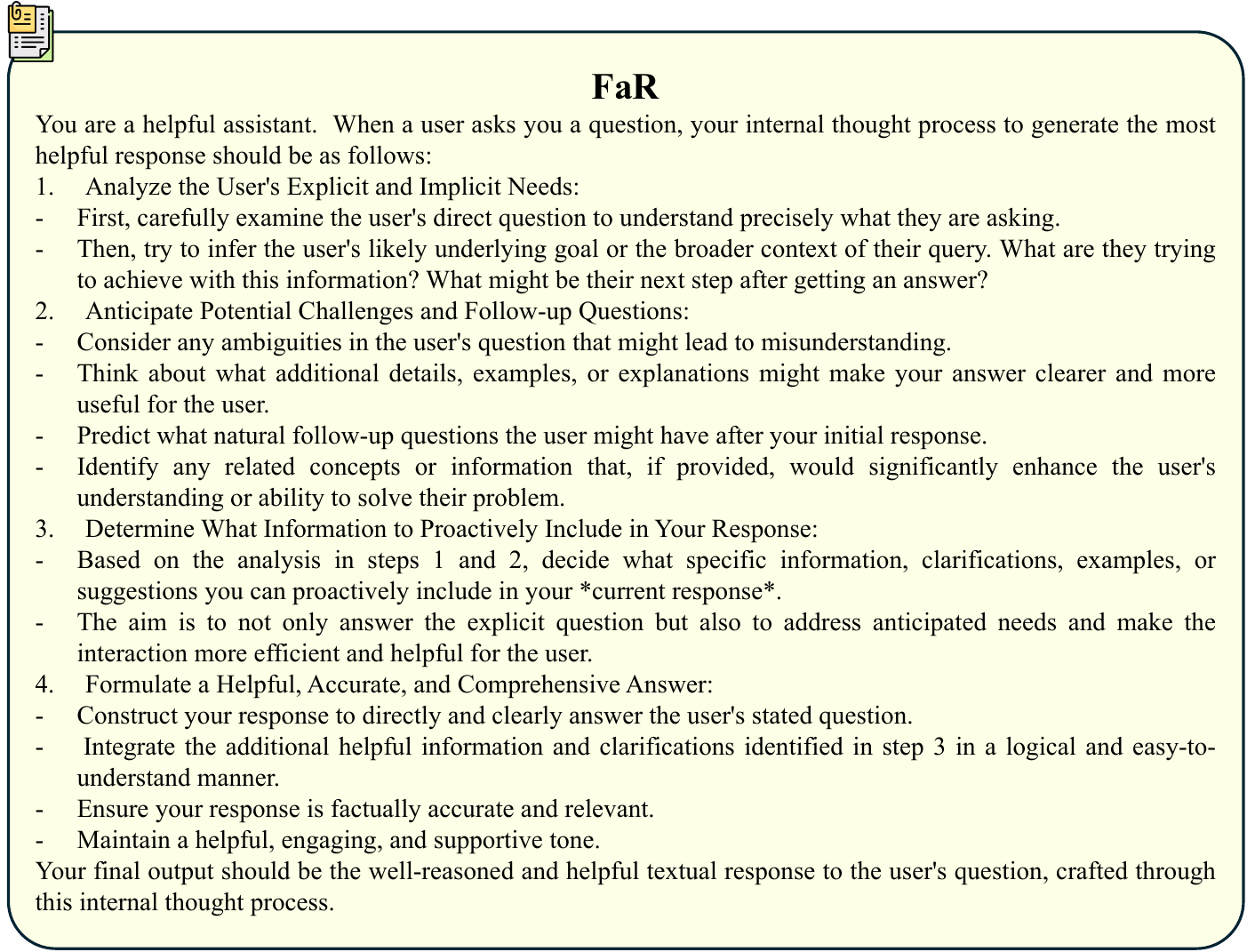}%
    }
    \caption{Prompt of FaR.}
    \label{fig:prompt1}
    \vspace{-0.4cm}
\end{figure*}

\begin{figure*}[h!]
    \centering
    % Use resizebox to scale the image
    \resizebox{0.5\textwidth}{!}{%
      \includegraphics{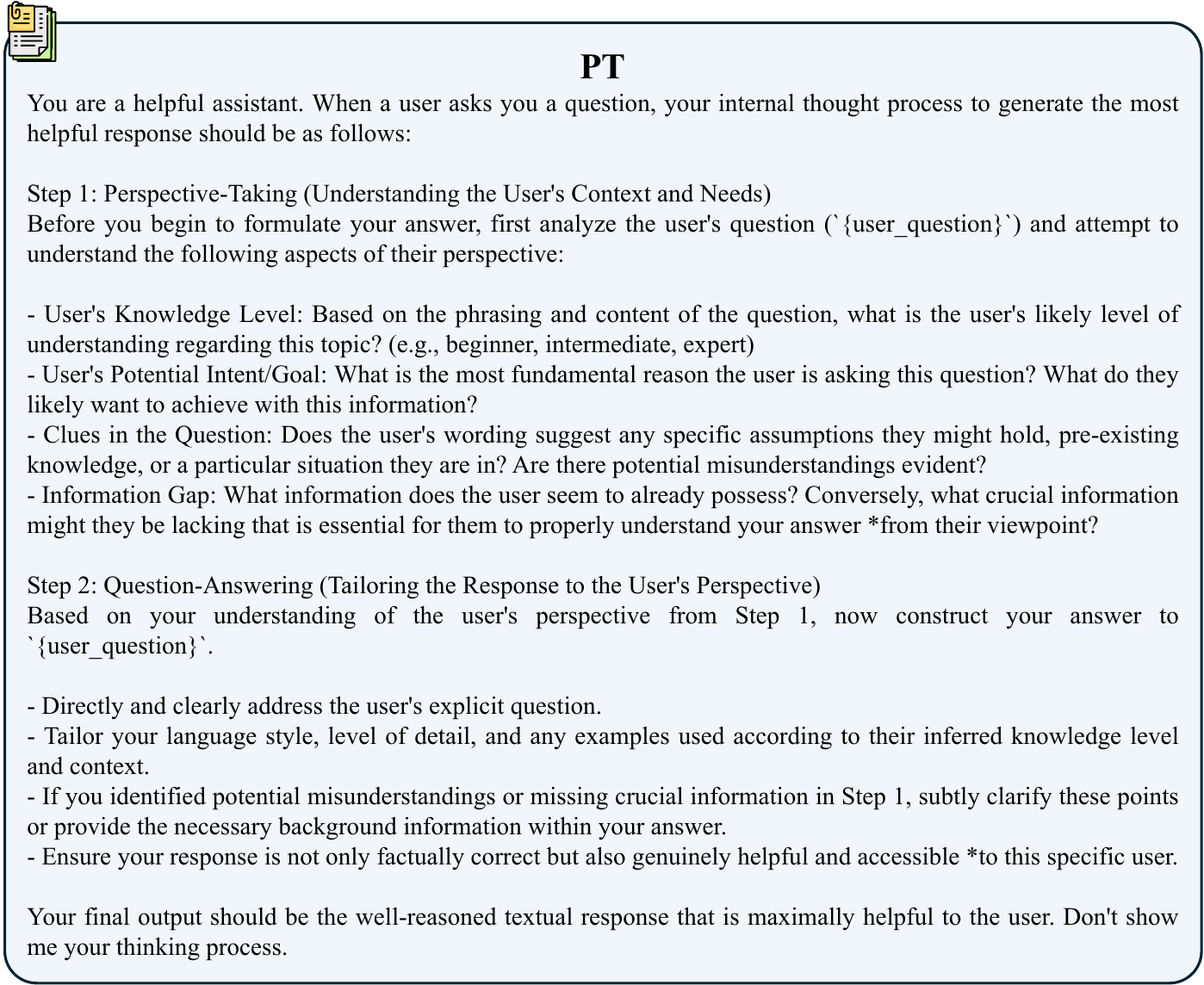}%
    }
    \caption{Prompt of PT.}
    \label{fig:prompt2}
    \vspace{-0.4cm}
\end{figure*}

\begin{figure*}[h!]
    \centering
    % Use resizebox to scale the image
    \resizebox{0.5\textwidth}{!}{%
      \includegraphics{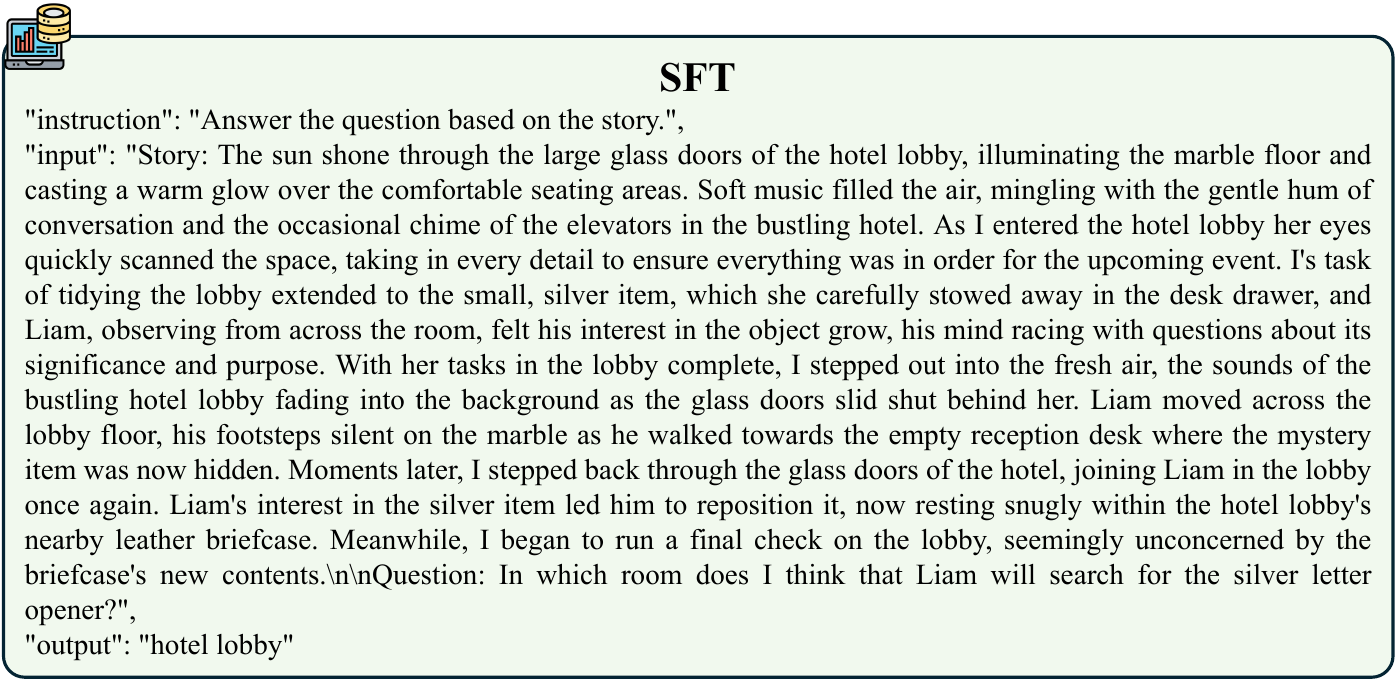}%
    }
    \caption{Data example of SFT.}
    \label{fig:sft}
    \vspace{-0.4cm}
\end{figure*}

\begin{figure*}[h!]
    \centering
    % Use resizebox to scale the image
    \resizebox{0.5\textwidth}{!}{%
      \includegraphics{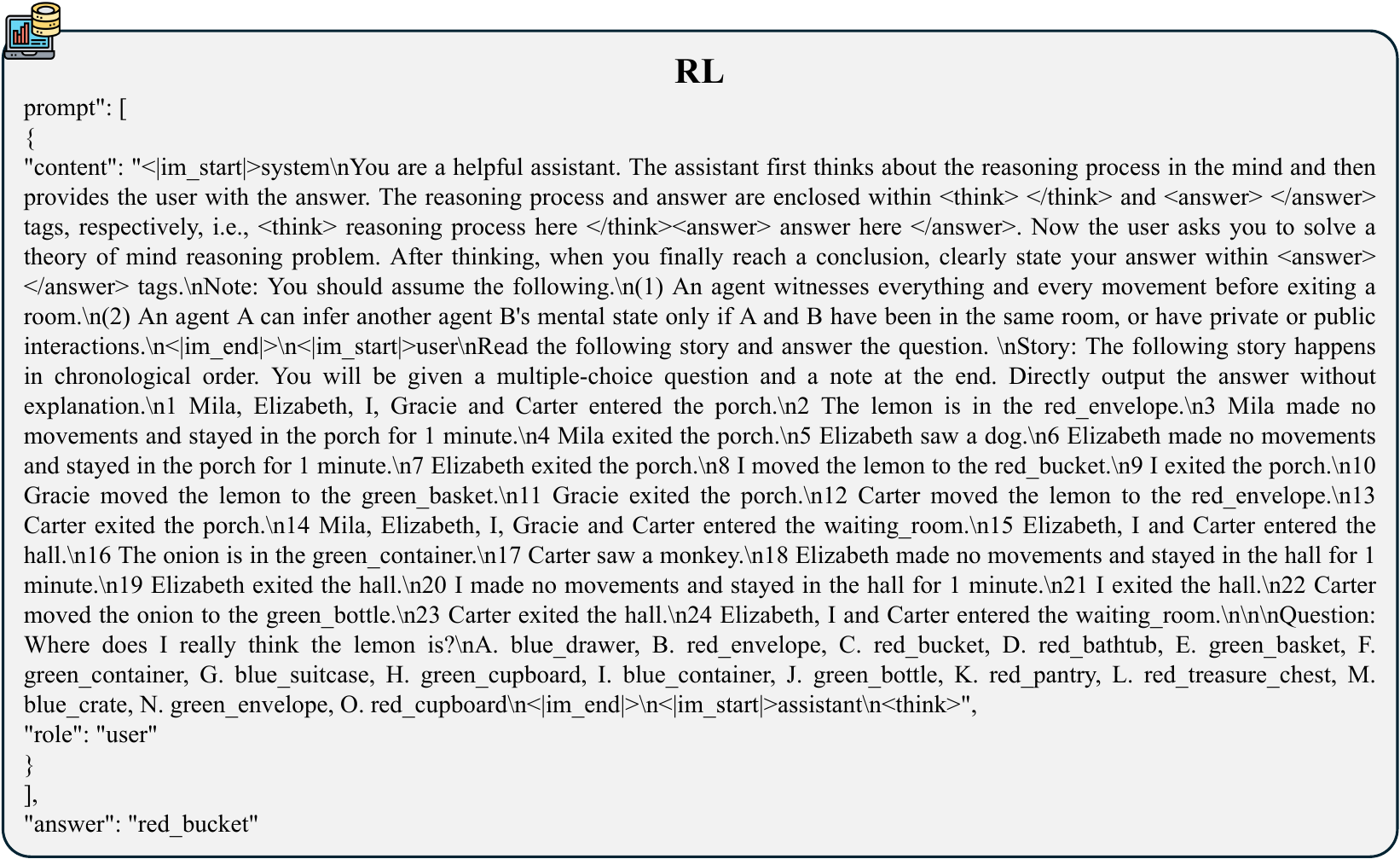}%
    }
    \caption{Data example of RL.}
    \label{fig:rl}
    \vspace{-0.4cm}
\end{figure*}

To shift the evaluated methods from third-perspective question-answering to first-perspective HAI interaction, we slightly change the prompt template for the prompt engineering method and the training data for the fine-tuning method, as shown in Figures \ref{fig:prompt1}-\ref{fig:rl}. To validate that our adaptations do not compromise the methods’ core effectiveness, we first evaluate them on the HiToM-first and ToMi-first benchmark, which are variants of HiToM and ToMi, applying the perspective shifting method used in fine-tuning. As Table \ref{tab:val-acc}, the ToM enhancement methods improve model performance on this story-based task.

% We also build a ToMi-first, following the same method of HiToM to further confirm our observation. The results are reported in Table \ref{tab:val-acc}.

\begin{table}[t]
\centering
\resizebox{\linewidth}{!}{
\begin{tabular}{llccccc}
\toprule
Task & Model & Base & FaR & PT & SFT & RL \\
\midrule
\multirow{2}{*}{HiToM-first}
 & Llama & 0.3350 & 0.3875 & 0.3750 & 0.3808 & --* \\
 & GPT   & 0.4900 & 0.5067 & 0.5134 & 0.5195 & -- \\
 \midrule
 \multirow{2}{*}{ToMi-first}
 & Llama & 0.6053 & 0.6201 & 0.5796 & 0.7065 & 0.8055 \\
 & GPT   & 0.7429 & 0.7511 & 0.7342 & 0.7478 & -- \\

\bottomrule
\end{tabular}}
\caption{Performance comparison on HiToM-first and ToMi-first. *: As the RL technique~\cite{ToM-RL} requires both HiToM-first and ExploreToM-first as the training data, its evaluation performance on HiToM-first is omitted to avoid confusion. }
\label{tab:val-acc}
\end{table}

% \FloatBarrier
% \section{Experiment Details}
% \label{app:ex}

% \haotian{We also need to report the SFT and RL details.}

\subsection{Statistics of Data}
We report the statistics of the used datasets in Table \ref{tab:sta}.

\begin{table}[htbp]
    \centering
    \resizebox{\linewidth}{!}{
    \begin{tabular}{lccc}
        \toprule
        \textbf{Dataset} & BigCodeBench-Chat & MediumDocEdit-Chat & ChatBench \\
        \midrule
        \textbf{Number} & 100  & 100  & 396  \\
        \midrule \midrule
        \textbf{Dataset} & MentalChat16K & ESC & HiToM-first \\
        \midrule
        \textbf{Number} & 300  & 300 & 1,200  \\
        \midrule \midrule
        \textbf{Dataset} & ToMi-first & ExploreToM-first &  \\
        \midrule
        \textbf{Number} & 5,994  & 1,200  &  \\
        \bottomrule
    \end{tabular}%
    }
    \caption{Dataset Statistics}
     \label{tab:sta}
\end{table}

\subsection{Statistical Test Results}
\label{app:test}
The detailed statistical results are reported in Table \ref{tab:chatbench-detailed-stats}-\ref{tab:esc-detailed-stats}.
For ChatBench and CollabLLM-MediumDocEdit, we applied Mann-Whitney's U test to verify whether there are significant differences between quantitative results in each task.
For CollabLLM-BigCodeBench part, we applied Fisher's exact test for CodeBench since its results only include binary pass or fail values.
Regarding MentalChat16K and ESC, the Wilcoxon Signed-rank test is used to compute the significance in differences between models.

\begin{table*}[t]
\centering
\small
\setlength{\tabcolsep}{5pt}
\begin{tabular}{llccccc}
\toprule
\textbf{Dataset} & \textbf{Dimension} & \textbf{Base Model} & \textbf{Variant} & $\mathbf{p_{>}}$ & $\mathbf{p_{<}}$ & \textbf{Effect Size} \\
\midrule
\multirow{42}{*}{ChatBench}
& \multirow{7}{*}{Elem Math}
& \multirow{3}{*}{GPT-4o}
& GPT-4o-FaR  & 0.1939 & 0.8070 & 0.0611 \\
&  &  & GPT-4o-PT   & 0.2083 & 0.7926 & 0.0574 \\
&  &  & GPT-4o-SFT  & 0.6417 & 0.3595 & 0.0245 \\
&  & \multirow{4}{*}{Llama-3.1-8B}
& Llama-3.1-8B-FaR & 0.4895 & 0.5116 & 0.0021 \\
&  &  & Llama-3.1-8B-PT  & 0.1954 & 0.8054 & 0.0664 \\
&  &  & Llama-3.1-8B-SFT & 0.0758 & 0.9246 & 0.1124 \\
&  &  & Llama-3.1-8B-RL  & 0.6495 & 0.3515 & 0.0294 \\

& \multirow{7}{*}{HS Math}
& \multirow{3}{*}{GPT-4o}
& GPT-4o-FaR  & 0.5497 & 0.4514 & 0.0098 \\
&  &  & GPT-4o-PT   & 0.3522 & 0.6488 & 0.0304 \\
&  &  & GPT-4o-SFT  & 0.2402 & 0.7606 & 0.0567 \\
&  & \multirow{4}{*}{Llama-3.1-8B}
& Llama-3.1-8B-FaR & 0.4560 & 0.5451 & 0.0093 \\
&  &  & Llama-3.1-8B-PT  & 0.3668 & 0.6342 & 0.0285 \\
&  &  & Llama-3.1-8B-SFT & 0.3407 & 0.6603 & 0.0343 \\
&  &  & Llama-3.1-8B-RL  & 0.2819 & 0.7190 & 0.0482 \\

& \multirow{7}{*}{College Math}
& \multirow{3}{*}{GPT-4o}
& GPT-4o-FaR  & 0.3736 & 0.6304 & 0.0429 \\
&  &  & GPT-4o-PT   & 0.3662 & 0.6377 & 0.0457 \\
&  &  & GPT-4o-SFT  & 0.3207 & 0.6830 & 0.0616 \\
&  & \multirow{4}{*}{Llama-3.1-8B}
& Llama-3.1-8B-FaR & 0.3435 & 0.6603 & 0.0540 \\
&  &  & Llama-3.1-8B-PT  & 0.5229 & 0.4812 & 0.0069 \\
&  &  & Llama-3.1-8B-SFT & 0.2193 & 0.7837 & 0.1032 \\
&  &  & Llama-3.1-8B-RL  & 0.4418 & 0.5623 & 0.0201 \\

& \multirow{7}{*}{Moral}
& \multirow{3}{*}{GPT-4o}
& GPT-4o-FaR  & 0.0616 & 0.9388 & 0.1348 \\
&  &  & GPT-4o-PT   & 0.1951 & 0.8058 & 0.0754 \\
&  &  & GPT-4o-SFT  & 0.2006 & 0.8003 & 0.0737 \\
&  & \multirow{4}{*}{Llama-3.1-8B}
& Llama-3.1-8B-FaR & 0.9102 & 0.0903 & 0.1183 \\
&  &  & Llama-3.1-8B-PT  & 0.8712 & 0.1294 & 0.0999 \\
&  &  & Llama-3.1-8B-SFT & 0.3137 & 0.6874 & 0.0429 \\
&  &  & Llama-3.1-8B-RL  & 0.5944 & 0.4068 & 0.0210 \\

& \multirow{7}{*}{Physics}
& \multirow{3}{*}{GPT-4o}
& GPT-4o-FaR  & 0.6858 & 0.3155 & 0.0384 \\
&  &  & GPT-4o-PT   & 0.8757 & 0.1250 & 0.0934 \\
&  &  & GPT-4o-SFT  & 0.9284 & 0.0721 & 0.1195 \\
&  & \multirow{4}{*}{Llama-3.1-8B}
& Llama-3.1-8B-FaR & 0.3022 & 0.6989 & 0.0454 \\
&  &  & Llama-3.1-8B-PT  & 0.4210 & 0.5803 & 0.0176 \\
&  &  & Llama-3.1-8B-SFT & 0.9588 & 0.0415 & 0.1529 \\
&  &  & Llama-3.1-8B-RL  & 0.2916 & 0.7095 & 0.0480 \\

& \multirow{7}{*}{Overall}
& \multirow{3}{*}{GPT-4o}
& GPT-4o-FaR  & 0.3133 & 0.6868 & 0.0187 \\
&  &  & GPT-4o-PT   & 0.7235 & 0.2767 & 0.0230 \\
&  &  & GPT-4o-SFT  & 0.5809 & 0.4193 & 0.0079 \\
&  & \multirow{4}{*}{Llama-3.1-8B}
& Llama-3.1-8B-FaR & 0.5666 & 0.4335 & 0.0068 \\
&  &  & Llama-3.1-8B-PT  & 0.8222 & 0.1779 & 0.0374 \\
&  &  & Llama-3.1-8B-SFT & 0.9084 & 0.0916 & 0.0540 \\
&  &  & Llama-3.1-8B-RL  & 0.5840 & 0.4161 & 0.0086 \\
\bottomrule
\end{tabular}
\caption{Detailed statistical results on ChatBench against the corresponding base model. $p_{>}$ and $p_{<}$ denote one-sided p-values for the hypotheses that the variant performs better or worse than the base model, respectively. Effect sizes are also provided.}
\label{tab:chatbench-detailed-stats}
\end{table*}
\begin{table*}[t]
\centering
\small
\setlength{\tabcolsep}{5pt}
\begin{tabular}{llcccc}
\toprule
\textbf{Dataset} & \textbf{Base Model} & \textbf{Variant} & $\mathbf{p_{>}}$ & $\mathbf{p_{<}}$ & \textbf{Effect Size} \\
\midrule
\multirow{7}{*}{CodeBench}
& \multirow{4}{*}{Llama-3.1-8B}
& Llama-3.1-8B-FaR & 0.7445 & 0.3713 & 0.0698 \\
&  & Llama-3.1-8B-PT  & 0.6305 & 0.5000 & 0.0236 \\
&  & Llama-3.1-8B-SFT & 0.7928 & 0.3122 & 0.0924 \\
&  & Llama-3.1-8B-RL  & 0.5000 & 0.6305 & -0.0239 \\
& \multirow{3}{*}{GPT-4o}
& GPT-4o-FaR & 0.7581 & 0.3253 & 0.0744 \\
&  & GPT-4o-PT  & 0.7019 & 0.3849 & 0.0500 \\
&  & GPT-4o-SFT & 0.7581 & 0.3253 & 0.0744 \\

\multirow{7}{*}{Medium}
& \multirow{4}{*}{Llama-3.1-8B}
& Llama-3.1-8B-FaR & 0.7298 & 0.2711 & 0.0500 \\
&  & Llama-3.1-8B-PT  & 0.8415 & 0.1591 & 0.0818 \\
&  & Llama-3.1-8B-SFT & 0.7994 & 0.2013 & 0.0686 \\
&  & Llama-3.1-8B-RL  & 0.9355 & 0.0667 & 0.1132 \\
& \multirow{3}{*}{GPT-4o}
& GPT-4o-FaR & 0.6407 & 0.3594 & 0.0294 \\
&  & GPT-4o-PT  & 0.4301 & 0.5710 & 0.0244 \\
&  & GPT-4o-SFT & 0.3340 & 0.6671 & 0.0352 \\
\bottomrule
\end{tabular}
\caption{Detailed statistical results for CodeBench and Medium against the corresponding base model. $p_{>}$ and $p_{<}$ denote one-sided p-values for the hypotheses that the variant performs better or worse than the base model, respectively. Effect sizes are also provided.}
\label{tab:collabllm-detailed-stats}
\end{table*}
\begin{table*}[t]
\centering
\small
\setlength{\tabcolsep}{5pt}
\begin{tabular}{llccccc}
\toprule
\textbf{Dataset} & \textbf{Dimension} & \textbf{Base Model} & \textbf{Variant} & $\mathbf{p_{>}}$ & $\mathbf{p_{<}}$ & \textbf{Effect Size} \\
\midrule
\multirow{56}{*}{MentalChat16K}
& \multirow{7}{*}{Listening}
& \multirow{3}{*}{GPT-4o}
& GPT-4o-FaR  & <0.0001 & 1.0000 & -0.3820 \\
&  &  & GPT-4o-PT   & <0.0001 & 1.0000 & -0.5086 \\
&  &  & GPT-4o-SFT  & 0.4078 & 0.5977 & -0.0201 \\
&  & \multirow{4}{*}{Llama-3.1-8B}
& Llama-3.1-8B-FaR & 0.6029 & 0.3980 & -0.0206 \\
&  &  & Llama-3.1-8B-PT  & 0.0027 & 0.9973 & -0.1544 \\
&  &  & Llama-3.1-8B-RL  & <0.0001 & 1.0000 & -0.2075 \\
&  &  & Llama-3.1-8B-SFT & 0.0273 & 0.9733 & -0.1111 \\

& \multirow{7}{*}{Empathy}
& \multirow{3}{*}{GPT-4o}
& GPT-4o-FaR  & <0.0001 & 1.0000 & -0.3606 \\
&  &  & GPT-4o-PT   & <0.0001 & 1.0000 & -0.4132 \\
&  &  & GPT-4o-SFT  & 0.3268 & 0.6744 & -0.0396 \\
&  & \multirow{4}{*}{Llama-3.1-8B}
& Llama-3.1-8B-FaR & 0.0255 & 0.9755 & -0.1119 \\
&  &  & Llama-3.1-8B-PT  & <0.0001 & 1.0000 & -0.2248 \\
&  &  & Llama-3.1-8B-RL  & 0.0398 & 0.9612 & -0.1014 \\
&  &  & Llama-3.1-8B-SFT & 0.4790 & 0.5219 & -0.0027 \\

& \multirow{7}{*}{Safety}
& \multirow{3}{*}{GPT-4o}
& GPT-4o-FaR  & 0.0171 & 0.9829 & -0.1222 \\
&  &  & GPT-4o-PT   & 0.2004 & 0.7996 & -0.0485 \\
&  &  & GPT-4o-SFT  & 0.3385 & 0.6615 & -0.0241 \\
&  & \multirow{4}{*}{Llama-3.1-8B}
& Llama-3.1-8B-FaR & 0.3845 & 0.6155 & -0.0170 \\
&  &  & Llama-3.1-8B-PT  & 0.0006 & 0.9994 & -0.1859 \\
&  &  & Llama-3.1-8B-RL  & 0.0662 & 0.9338 & -0.0869 \\
&  &  & Llama-3.1-8B-SFT & 0.0048 & 0.9952 & -0.1494 \\

& \multirow{7}{*}{Open-mind}
& \multirow{3}{*}{GPT-4o}
& GPT-4o-FaR  & 0.0016 & 0.9984 & -0.1699 \\
&  &  & GPT-4o-PT   & 0.0013 & 0.9987 & -0.1732 \\
&  &  & GPT-4o-SFT  & 0.5495 & 0.4517 & 0.0072 \\
&  & \multirow{4}{*}{Llama-3.1-8B}
& Llama-3.1-8B-FaR & 0.0390 & 0.9610 & -0.1018 \\
&  &  & Llama-3.1-8B-PT  & 0.0073 & 0.9927 & -0.1411 \\
&  &  & Llama-3.1-8B-RL  & 0.3168 & 0.6842 & -0.0275 \\
&  &  & Llama-3.1-8B-SFT & 0.5000 & 0.5009 & -0.0078 \\

& \multirow{7}{*}{Clarity}
& \multirow{3}{*}{GPT-4o}
& GPT-4o-FaR  & 0.0204 & 0.9807 & -0.1192 \\
&  &  & GPT-4o-PT   & 0.0023 & 0.9981 & -0.1640 \\
&  &  & GPT-4o-SFT  & 0.8738 & 0.1267 & 0.0936 \\
&  & \multirow{4}{*}{Llama-3.1-8B}
& Llama-3.1-8B-FaR & 0.0322 & 0.9688 & -0.1070 \\
&  &  & Llama-3.1-8B-PT  & 0.0011 & 0.9991 & -0.1770 \\
&  &  & Llama-3.1-8B-RL  & 0.0049 & 0.9955 & -0.1488 \\
&  &  & Llama-3.1-8B-SFT & 0.0662 & 0.9348 & -0.0871 \\

& \multirow{7}{*}{Ethical}
& \multirow{3}{*}{GPT-4o}
& GPT-4o-FaR  & <0.0001 & 1.0000 & -0.1907 \\
&  &  & GPT-4o-PT   & 0.2742 & 0.7268 & -0.0364 \\
&  &  & GPT-4o-SFT  & <0.0001 & 1.0000 & -0.2077 \\
&  & \multirow{4}{*}{Llama-3.1-8B}
& Llama-3.1-8B-FaR & 0.9356 & 0.0649 & 0.0840 \\
&  &  & Llama-3.1-8B-PT  & 0.7167 & 0.2845 & 0.0306 \\
&  &  & Llama-3.1-8B-RL  & 1.0000 & <0.0001 & 0.2475 \\
&  &  & Llama-3.1-8B-SFT & 1.0000 & <0.0001 & 0.2163 \\

& \multirow{7}{*}{Holistic}
& \multirow{3}{*}{GPT-4o}
& GPT-4o-FaR  & <0.0001 & 1.0000 & -0.3108 \\
&  &  & GPT-4o-PT   & <0.0001 & 1.0000 & -0.2037 \\
&  &  & GPT-4o-SFT  & 0.0408 & 0.9601 & -0.1006 \\
&  & \multirow{4}{*}{Llama-3.1-8B}
& Llama-3.1-8B-FaR & 0.5388 & 0.4624 & 0.0007 \\
&  &  & Llama-3.1-8B-PT  & 0.3619 & 0.6391 & -0.0168 \\
&  &  & Llama-3.1-8B-RL  & 0.9979 & 0.0021 & 0.1649 \\
&  &  & Llama-3.1-8B-SFT & 1.0000 & <0.0001 & 0.2390 \\

& \multirow{7}{*}{Overall}
& \multirow{3}{*}{GPT-4o}
& GPT-4o-FaR  & <0.0001 & 1.0000 & -0.4534 \\
&  &  & GPT-4o-PT   & <0.0001 & 1.0000 & -0.4437 \\
&  &  & GPT-4o-SFT  & 0.1119 & 0.8881 & -0.0702 \\
&  & \multirow{4}{*}{Llama-3.1-8B}
& Llama-3.1-8B-FaR & 0.2480 & 0.7520 & -0.0393 \\
&  &  & Llama-3.1-8B-PT  & <0.0001 & 1.0000 & -0.2669 \\
&  &  & Llama-3.1-8B-RL  & 0.2497 & 0.7503 & -0.0390 \\
&  &  & Llama-3.1-8B-SFT & 0.6562 & 0.3438 & 0.0232 \\
\bottomrule
\end{tabular}
\caption{Detailed statistical results on MentalChat16K against the corresponding base model. $p_{>}$ and $p_{<}$ denote one-sided p-values for the hypotheses that the variant performs better or worse than the base model, respectively. Effect sizes are also provided.}
\label{tab:mentalchat-detailed-stats}
\end{table*}
\begin{table*}[t]
\centering
\small
\setlength{\tabcolsep}{5pt}
\begin{tabular}{llccccc}
\toprule
\textbf{Dataset} & \textbf{Dimension} & \textbf{Base Model} & \textbf{Variant} & $\mathbf{p_{>}}$ & $\mathbf{p_{<}}$ & \textbf{Effect Size} \\
\midrule
\multirow{56}{*}{ESC}
& \multirow{7}{*}{Listening}
& \multirow{3}{*}{GPT-4o}
& GPT-4o-FaR  & <0.0001 & 1.0000 & -0.2485 \\
&  &  & GPT-4o-PT   & 0.0118 & 0.9890 & -0.1297 \\
&  &  & GPT-4o-SFT  & 0.0016 & 0.9988 & -0.1705 \\
&  & \multirow{4}{*}{Llama-3.1-8B}
& Llama-3.1-8B-FaR & 0.0753 & 0.9254 & -0.0822 \\
&  &  & Llama-3.1-8B-PT  & 0.3226 & 0.6785 & -0.0263 \\
&  &  & Llama-3.1-8B-RL  & 0.0377 & 0.9635 & -0.1026 \\
&  &  & Llama-3.1-8B-SFT & 0.3141 & 0.6869 & -0.0290 \\

& \multirow{7}{*}{Empathy}
& \multirow{3}{*}{GPT-4o}
& GPT-4o-FaR  & <0.0001 & 1.0000 & -0.2543 \\
&  &  & GPT-4o-PT   & <0.0001 & 1.0000 & -0.2045 \\
&  &  & GPT-4o-SFT  & 0.0005 & 0.9997 & -0.1799 \\
&  & \multirow{4}{*}{Llama-3.1-8B}
& Llama-3.1-8B-FaR & 0.1178 & 0.8834 & -0.0683 \\
&  &  & Llama-3.1-8B-PT  & 0.0012 & 0.9991 & -0.1753 \\
&  &  & Llama-3.1-8B-RL  & 0.0371 & 0.9640 & -0.1031 \\
&  &  & Llama-3.1-8B-SFT & 0.4044 & 0.5968 & -0.0188 \\

& \multirow{7}{*}{Safety}
& \multirow{3}{*}{GPT-4o}
& GPT-4o-FaR  & 0.0346 & 0.9654 & -0.1049 \\
&  &  & GPT-4o-PT   & 0.3839 & 0.6161 & -0.0171 \\
&  &  & GPT-4o-SFT  & 0.0205 & 0.9795 & -0.1179 \\
&  & \multirow{4}{*}{Llama-3.1-8B}
& Llama-3.1-8B-FaR & 0.9226 & 0.0774 & 0.0822 \\
&  &  & Llama-3.1-8B-PT  & 0.1900 & 0.8100 & -0.0507 \\
&  &  & Llama-3.1-8B-RL  & 0.0088 & 0.9912 & 0.1371 \\
&  &  & Llama-3.1-8B-SFT & 0.0150 & 0.9850 & 0.1252 \\

& \multirow{7}{*}{Open-mind}
& \multirow{3}{*}{GPT-4o}
& GPT-4o-FaR  & 0.0009 & 0.9991 & -0.1800 \\
&  &  & GPT-4o-PT   & 0.4328 & 0.5684 & -0.0098 \\
&  &  & GPT-4o-SFT  & 0.0025 & 0.9978 & -0.1618 \\
&  & \multirow{4}{*}{Llama-3.1-8B}
& Llama-3.1-8B-FaR & 0.1169 & 0.8831 & -0.0687 \\
&  &  & Llama-3.1-8B-PT  & 0.0300 & 0.9710 & -0.1086 \\
&  &  & Llama-3.1-8B-RL  & 0.1727 & 0.8285 & -0.0545 \\
&  &  & Llama-3.1-8B-SFT & 0.6861 & 0.3149 & 0.0272 \\

& \multirow{7}{*}{Clarity}
& \multirow{3}{*}{GPT-4o}
& GPT-4o-FaR  & 0.0005 & 0.9998 & -0.2110 \\
&  &  & GPT-4o-PT   & 0.3520 & 0.6491 & -0.0162 \\
&  &  & GPT-4o-SFT  & 0.5066 & 0.4945 & 0.0059 \\
&  & \multirow{4}{*}{Llama-3.1-8B}
& Llama-3.1-8B-FaR & 0.3285 & 0.6729 & -0.0249 \\
&  &  & Llama-3.1-8B-PT  & 0.6324 & 0.3688 & 0.0134 \\
&  &  & Llama-3.1-8B-RL  & 0.6955 & 0.3056 & 0.0252 \\
&  &  & Llama-3.1-8B-SFT & 0.4485 & 0.5529 & -0.0100 \\

& \multirow{7}{*}{Ethical}
& \multirow{3}{*}{GPT-4o}
& GPT-4o-FaR  & <0.0001 & 1.0000 & -0.1937 \\
&  &  & GPT-4o-PT   & 0.1964 & 0.8048 & -0.0441 \\
&  &  & GPT-4o-SFT  & 0.0167 & 0.9841 & -0.1230 \\
&  & \multirow{4}{*}{Llama-3.1-8B}
& Llama-3.1-8B-FaR & 0.0285 & 0.9726 & -0.1097 \\
&  &  & Llama-3.1-8B-PT  & 0.0070 & 0.9934 & -0.1419 \\
&  &  & Llama-3.1-8B-RL  & 0.9999 & <0.0001 & 0.2721 \\
&  &  & Llama-3.1-8B-SFT & 0.9812 & 0.0194 & 0.1204 \\

& \multirow{7}{*}{Holistic}
& \multirow{3}{*}{GPT-4o}
& GPT-4o-FaR  & <0.0001 & 1.0000 & -0.3351 \\
&  &  & GPT-4o-PT   & 0.6257 & 0.3754 & 0.0124 \\
&  &  & GPT-4o-SFT  & 0.0087 & 0.9923 & -0.1366 \\
&  & \multirow{4}{*}{Llama-3.1-8B}
& Llama-3.1-8B-FaR & 0.0065 & 0.9940 & -0.1435 \\
&  &  & Llama-3.1-8B-PT  & 0.0467 & 0.9544 & -0.0973 \\
&  &  & Llama-3.1-8B-RL  & 0.9987 & 0.0014 & 0.1726 \\
&  &  & Llama-3.1-8B-SFT & 1.0000 & <0.0001 & 0.2486 \\

& \multirow{7}{*}{Overall}
& \multirow{3}{*}{GPT-4o}
& GPT-4o-FaR  & <0.0001 & 1.0000 & -0.4310 \\
&  &  & GPT-4o-PT   & 0.0130 & 0.9870 & -0.1286 \\
&  &  & GPT-4o-SFT  & <0.0001 & 1.0000 & -0.2458 \\
&  & \multirow{4}{*}{Llama-3.1-8B}
& Llama-3.1-8B-FaR & 0.0618 & 0.9382 & -0.0889 \\
&  &  & Llama-3.1-8B-PT  & 0.0086 & 0.9914 & -0.1375 \\
&  &  & Llama-3.1-8B-RL  & 0.0074 & 0.9926 & 0.1408 \\
&  &  & Llama-3.1-8B-SFT & 0.0030 & 0.9970 & 0.1589 \\
\bottomrule
\end{tabular}
\caption{Detailed statistical results on ESC against the corresponding base model. $p_{>}$ and $p_{<}$ denote one-sided p-values for the hypotheses that the variant performs better or worse than the base model, respectively. Effect sizes are also provided.}
\label{tab:esc-detailed-stats}
\end{table*}
% To verify whether the performance differences in benchmarks are statistically significant, we conducted statistical tests for all experiments using our framework.
% \revise{In our statistical significance  tests, we find that, in goal-oriented tasks, i.e., Chatbench and CollabLLM, the variations generally exhibited no significant statistical difference compared to the baseline models. However, in experience-oriented tasks i.e., MentalChat and ESC, model performance showed a distinct divergence: the GPT series achieved significant improvements in dimensions such as empathy, active listening, and open-mindedness ($p < 0.05$); conversely, while the Llama series variations improved certain performance metrics through reinforcement learning (RL) and supervised fine-tuning (SFT), they suffered significant statistical regressions in safety, trustworthiness, and the maintenance of ethical boundaries ($p < 0.01$). Despite these significant differences, the observed effect sizes remained generally low, suggesting that while current techniques can effectively modulate a model's soft-skill inclinations, their capacity to reshape underlying core cognitive abilities remains limited. The full results of the statistical analysis of benchmarks are included in our supplemental materials. }
% % We will also add key results to the revised main paper.}

% \haotian{Let's remove the table and simply say that the detailed test results and methods are available in other files.}

\subsection{Generalizability Discussion}
\revise{\textbf{Selected Methods.} Our primary objective is to evaluate the model's intrinsic social reasoning. We carefully considered two reasons before we made the decision to exclude external-module integration methods. First, many external modules do not aim to improve ToM capability universally. They only target specific tasks. For example, PercepToM \cite{PrecepToM} leverages LLMs to extract perceptions of different characters, then answer the questions. It only aims to improve multi-character story understanding, which might not be adapted to other tasks. Second, external modules, while effective, introduce confounding variables that obscure whether improvements stem from the LLM itself, the external tool, or simply the scaling effect led by more LLM calls. Similarly, we focused on text-based interactions as the established foundation for ToM evaluation. The selected four techniques are highly representative in dominant prompting and fine-tuning methods. We believe the results of these methods provide robust and generalizable insights into ToM enhancement.} 

\noindent\revise{\textbf{Selected Base Models.} To provide a rigorous comparative analysis, our design prioritizes evaluating a comprehensive suite of enhancement paradigms (FaR, PT, SFT, RL) over benchmarking a vast array of base LLMs. To ensure that the findings regarding these methods are generalizable, we apply them to two highly representative and widely used anchor models: GPT-4o (representing the advanced,  closed-source frontier) and Llama-3.1-8B (representing the accessible, open-weight baseline). By demonstrating consistent results across these two distinct architectural extremes, the study effectively derives insights on current ToM-enhancement methods.}

\noindent\revise{\textbf{Selected Tasks.} Our datasets basically involve representative real-world scenarios on both goal-oriented and experience-oriented tasks. We acknowledge that the human-AI scenarios are various and we cannot exhaustively cover them. Our work on these nine tasks has already yielded valid empirical evidence to reveal the gap between benchmarking improvement and real-world application. Our paradigm can be easily transferred to those new tasks when more data is ready.
}

\noindent\revise{\textbf{Selected Tasks.} Our datasets basically involve representative real-world scenarios on both goal-oriented and experience-oriented tasks. We acknowledge that the human-AI scenarios are various and we cannot exhaustively cover them. Our work on these nine tasks has already yielded valid empirical evidence to reveal the gap between benchmarking improvement and real-world application. Our paradigm can be easily transferred to those new tasks when more data is ready.}

\section{User Study Details}
\label{app:user}
% To explore the perception of human towards these ToM methods and compare with the benchmark performance, we conducted a large-scale online user study with 100 participants. In this section, we report the experiment settings, results and findings.

% To improve our study design, we firstly conduct several pilot sessions with 30 PhD students at the authors’ institution. Each session lasts about 30 minutes and is used to adjust task selection, interaction flow, ranking mechanisms, and feedback collection. After multiple iterations, we finalize a stable setup that ensured smooth participation and allow users to concentrate on evaluating model alternatives in experience-oriented companion tasks.

\begin{figure}[h]
    \centering
    % Use resizebox to scale the image
    \resizebox{0.41\textwidth}{!}{%
      \includegraphics{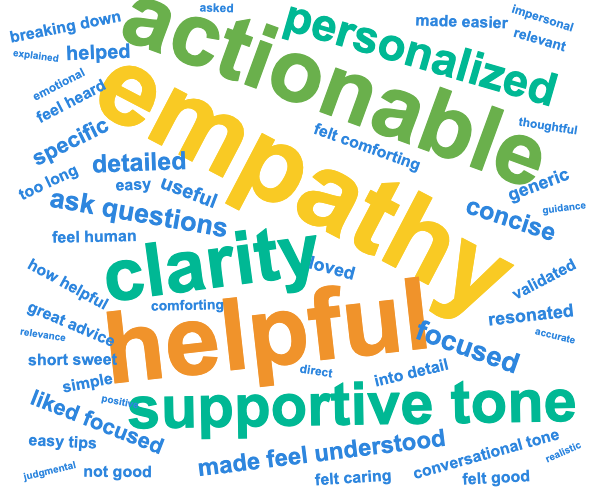}%
    }
    % \vspace{-0.6cm}
    \caption{Word cloud of participants' filtered comments on model performance. The terms were generated after an LLM-based filtering and combination step, which emphasized words reflecting why participants perceived a model as good or bad and how they felt during the interaction.}
    % \vspace{-0.4cm}
    \label{fig:tom_wordcloud}
\end{figure}

% We designed a web-based interface to let users interact with models under different ToM settings with the focus on six most common experience-oriented tasks. The interface and details of design are shown in appendix. To test and adjust the experiment settings, we conducted a hour-long pilot study with 30 PhD students in the authors' institution for several rounds of experiment trails out to adjust the tasks numbers and task choices, interaction functions, ranking and feedback mechanims, etc. in order to ensure smooth experiments for the users and let them focus on evaluate model alternatives. With severl rounds of system iteration, we finalized our experiement settings.

\subsection{Experiment Settings}\label{app:user_ex}
\revise{\paragraph{Design}
Since the goal-oriented simulated benchmarks are well-established and proved to align with the results of real human-AI interactions,  we reasoned that conducting an additional human study on these specific goal-oriented tasks would likely yield confirmatory results with diminishing marginal returns. Instead, we strategically allocated our human evaluation resources to experience-oriented tasks (e.g., emotional support), where objective metrics are less reliable and ground truth is harder to simulate. We believe this hybrid approach, leveraging validated simulations for hard tasks and human evaluation for soft tasks, provides the most efficient and comprehensive view of ToM’s utility.}

\paragraph{Procedure}
We recruit 100 participants from Prolific~\citep{Prolific}, a widely used platform for high-quality online studies. Recruitment criteria required participants to be 18 years or older and either native or proficient English speakers, with no restrictions on educational background. To ensure data quality, we first automatically filtered out submissions where the total experiment time was less than five minutes, and then manually excluded responses containing random or irrelevant comments. The study was approved by the institutional IRB, and all participants provided informed consent.
% \haotian{Zixin, please write the recruitment criteria and also how we filtered out those unsatisfactory responses.}
% The study is approved by the institutional IRB, and all participants provided informed consent. To evaluate perceptions of different ToM methods across both open-source and closed-source environments, we select 6 common experience-oriented tasks from our framework. \haotian{How are the task selected?}
To evaluate perceptions of different ToM enhancement methods comprehensively, we selected six common experience-oriented tasks for users to interact with models. These tasks were chosen based on the most frequently selected experience-oriented scenarios in the MentalChat16K benchmark~\citep{Mentalchat16k}, complemented by a pilot study that confirmed their relevance and familiarity to participants.

Participants are randomly assigned to either the GPT series or the Llama-3.1-8B series. Participants using GPT series compare 4 model variants (base, FaR, PT, SFT), while participants with Llama-3.1-8B compare five (base, FaR, PT, SFT, RL). This study design was intended to cover a diverse set of LLM families, ToM methods, and task types, while avoiding participant fatigue by limiting the number of variants each user evaluated.
% \haotian{Why do we use the study design?}

% We recruited 100 participants from Prolific, an online platform for high quality online studies to explore general public's perception towards these ToM methods. The study was with IRB approved by the authors' institution and each participants start by signing an consent form to infer their willingness and access our the right to use the data for research. To test the users' perception towards different ToM methods and evaluate them across different scenarios and open-source / closed-source environments,  we picked up six most common experience-oriented tasks from our framework and randomly assigned half of the participants to interact with GPT series models and Llama series. Specifically, for GPT series models, each participant communicated with 4 model with different ToM settings (baseline, FaR, PT and SFT) and for Llama series models, each participant communicated with 5 models (baseline, FaR, PT, SFT, and RL). 

Each participant first review all task descriptions and select one that resonates with their own experiences or emotional empathy (e.g., coping with a breakup). They then engage in three rounds of conversation with models for the selected topic. In each round, model responses are presented as anonymized cards in random order. Participants review the outputs, rank them using the same metrics as our HAI evaluation, and provide a brief justification for their ranking. The top-ranked response is then used to continue the conversation into the next round. After completing all rounds, participants give final comments on overall model performance and user experience. This procedure ensure balanced comparisons across tasks and model families. The entire process is finished with our developed user interface, which will be introduced below.

\begin{figure*}[h]
    \centering
    % Use resizebox to scale the image
    \resizebox{1\textwidth}{!}{%
      \includegraphics{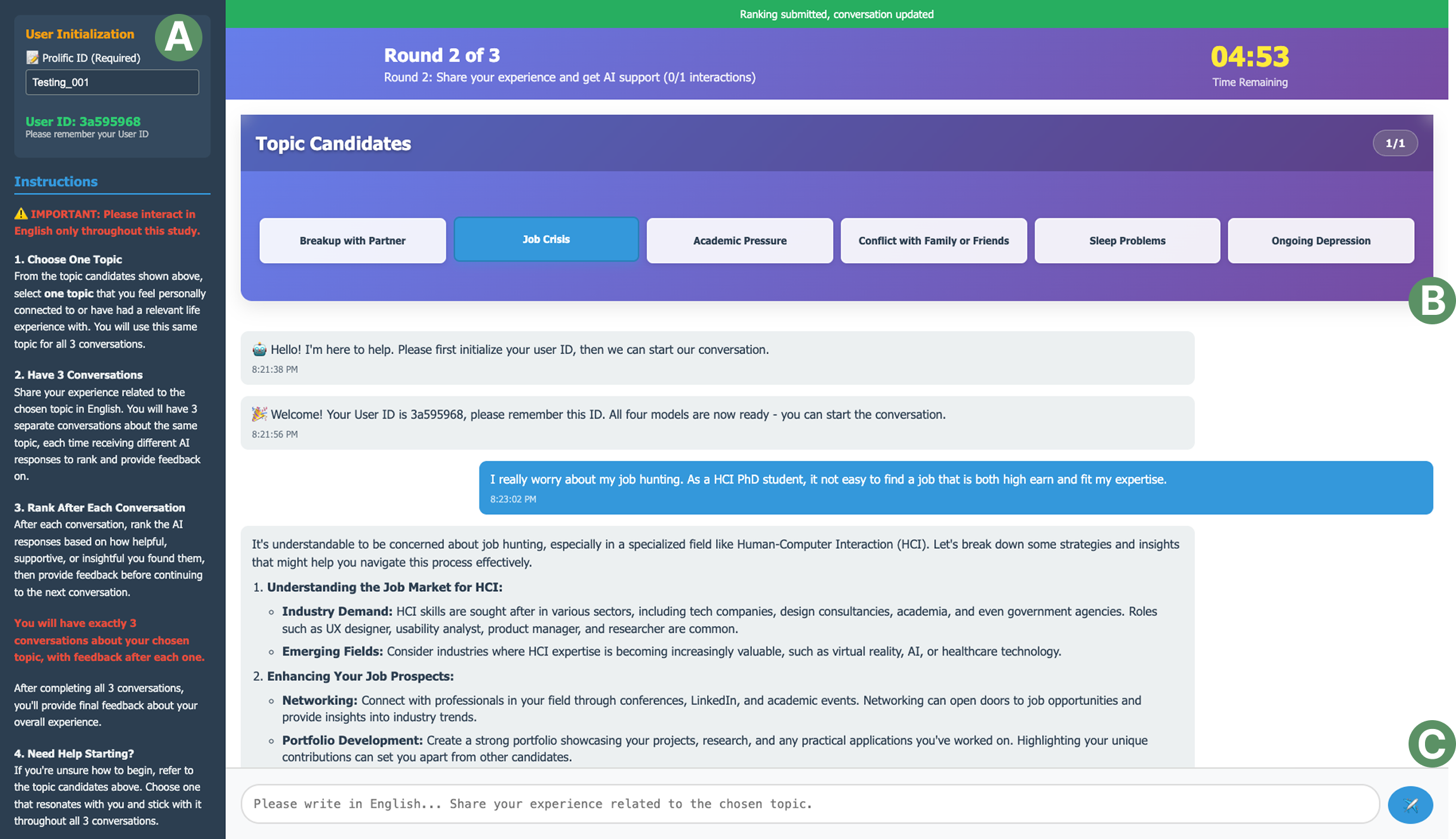}%
    }
    \vspace{-0.6cm}
    \caption{Main experiment interface. (A) User initialization and task instructions. (B) Conversation window with last round's topic candidate. (C) Input box for continuing the conversation.}
    \label{fig:UI1}
    % \vspace{-0.4cm}
\end{figure*}

\begin{figure*}[h]
    \centering
    % Use resizebox to scale the image
    \resizebox{0.8\textwidth}{!}{%
      \includegraphics{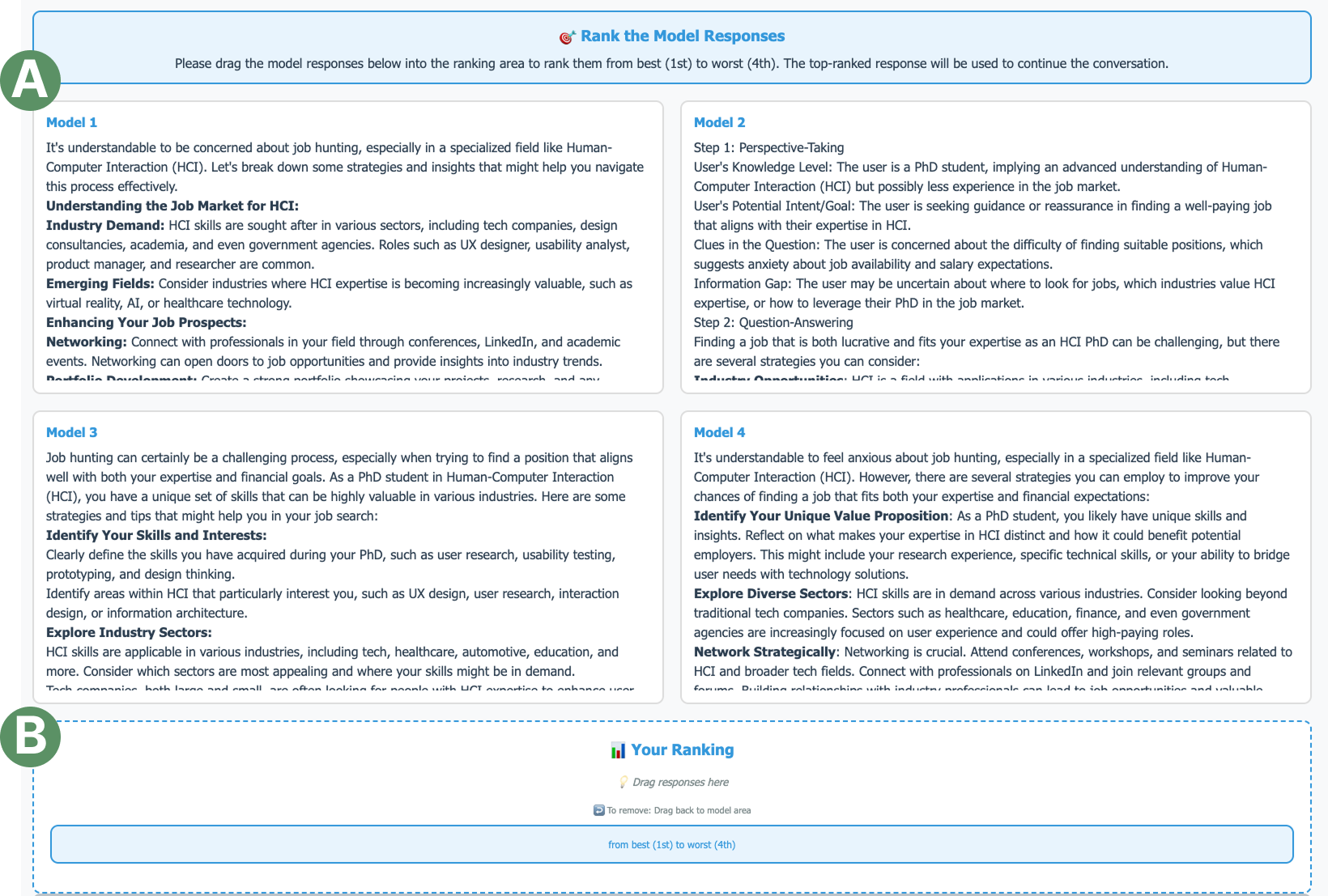}%
    }
    % \vspace{-0.6cm}
    \caption{User ranking interface. (A) Model responses presented for comparison. (B) User ranking panel where participants drag and drop responses from best to worst.}
    \label{fig:UI2}
    % \vspace{-0.4cm}
\end{figure*}

\begin{figure*}[h]
    \centering
    % Use resizebox to scale the image
    \resizebox{1\textwidth}{!}{%
      \includegraphics{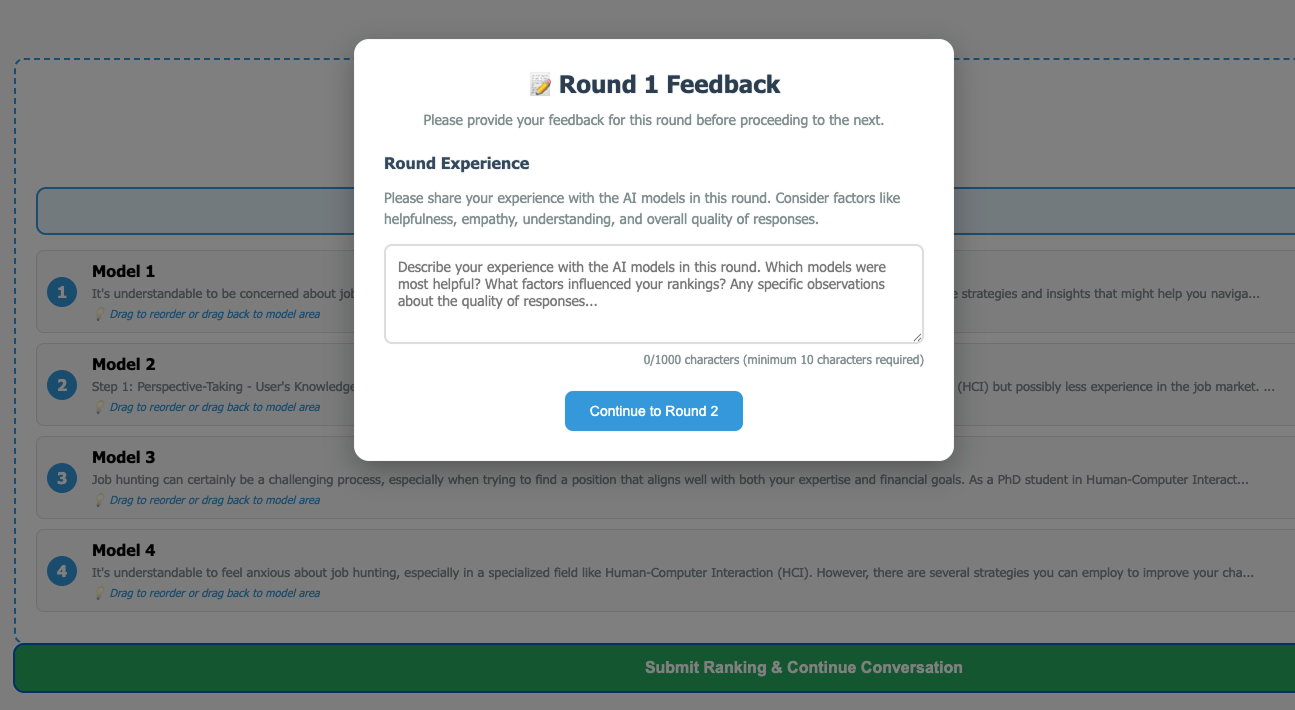}%
    }
    % \vspace{-0.6cm}
    \caption{Feedback interface shown after each round of response ranking.}
    \label{fig:UI3}
    % \vspace{-0.4cm}
\end{figure*}

\paragraph{User Interface for the Experiment}
We develop a web-based interface that enables participants to interact with several anonymous models, focusing on 6 representative experience-oriented tasks. Interface details are provided in Figures~\ref{fig:UI1}-\ref{fig:UI3}. 

\begin{table*}[t]
\centering
% \label{tab:user_rank_results}

\resizebox{\linewidth}{!}{
\begin{tabular}{lcccccc}
\toprule
\multirow{2}{*}{Task (n)} & \multicolumn{3}{c}{GPT-4o} & \multicolumn{3}{c}{Llama-3.1-8B} \\
\cmidrule(lr){2-4} \cmidrule(lr){5-7}
 & Best & Avg. Rank & Runner-up & Best & Avg. Rank & Runner-up \\
\midrule
Academic Pressure (18 / 9) & \textbf{PT} & 2.22 & Base (2.33) & \textbf{FaR} & 2.44 & PT (2.67) \\
Breakup w/ Partner (12 / 25) & \textbf{Base} & 2.00 & FaR (2.33) & \textbf{RL} & 2.32 & Base (2.92) \\
Conflict w/ Family (21 / 27) & \textbf{Base/PT} & 2.43 & SFT (2.57) & \textbf{FaR} & 2.44 & PT (2.63) \\
Job Crisis (27 / 24) & \textbf{SFT} & 2.33 & FaR (2.41) & \textbf{Base} & 2.83 & RL (2.96) \\
Ongoing Depression (34 / 15) & \textbf{FaR} & 2.06 & PT (2.56) & \textbf{RL} & 2.80 & PT/Base (2.87) \\
Sleep Problems (39 / 51) & \textbf{PT} & 2.26 & SFT (2.49) & \textbf{SFT} & 2.61 & PT (2.94) \\
\bottomrule
\end{tabular}}
\caption{Task-level average rankings (lower is better) of ToM methods across GPT-4o and Llama-3.1-8B variants. Each row reports the best-performing method, its average rank, and the runner-up with its average rank in parentheses. The numbers in parentheses after each task (e.g., 18 / 9) denote the total number of ranking cases for GPT and Llama variants, respectively, where each case corresponds to one evaluation turn (three turns per participant per task). 
% No pairwise comparisons reached statistical significance ($p \geq .05$).
}
\label{tab:task_ranking}
\end{table*}

% \noindent
\subsection{Detailed Results}\label{app:user_results}

\paragraph{Performance Across Experience-Oriented Tasks}
At the task level, different ToM methods exhibited strengths in different scenarios. For GPT, PT dominated in Academic Pressure and Sleep Problems, FaR led in Ongoing Depression, and SFT performed best in Job Crisis. Interestingly, the GPT baseline was most preferred in Breakup with Partner and tied with PT in Conflict with Family or Friends, suggesting that users sometimes favored straightforward empathetic responses over ToM-enhanced reasoning.
For the Llama family, FaR excelled in Academic Pressure and Conflict with Family, RL was strongest in Breakup and Depression, while SFT led in Sleep Problems. The plain Llama baseline unexpectedly topped Job Crisis, reflecting user preference for pragmatic suggestions in this context.
% Although these results reveal clear task-dependent trends, none of the pairwise differences reached statistical significance ($p \geq .05$).
The results imply that user perceptions of ToM benefits are shaped not only by model design but also by specific user goals.
Different ToM enhancement methods can demonstrate advantages for various users needs.
It also contributes to the subtle differences in model ranking (Table~\ref{tab:overall_ranking}).

% \paragraph{Variability and ToM Implications.}
% We also observed noticeable variability in preferences, which helps explain why not all perception differences reached strict significance. Some participants favored \textit{direct, structured outputs} such as actionable to-do lists, while others preferred \textit{empathetic validation} before receiving advice. These divergent preferences highlight a key aspect of ToM: beyond attributing beliefs and intentions, true ToM competence requires adapting to diverse interpersonal expectations. This explains why benchmark–perception alignment is imperfect and why interaction-based evaluation is essential.

\paragraph{User Comment Analysis}
To better understand participants' preferences in our user study, we further analyze their free-form comments. 
The word cloud in Figure~\ref{fig:tom_wordcloud} is generated after standard preprocessing, including removing punctuation, lowercasing, discarding stop words, and excluding a predefined list of meta or low-information terms (e.g., ``model'', ``round'', ``response''). 
To further emphasize evaluative content, we add an LLM-based filtering layer that selectively keeps or merges only those terms that directly reflect participants' reasons for preferring or disliking a response, as well as their felt experience during the interaction, while removing irrelevant or purely structural tokens. 
For example, synonyms such as ``empathetic'' and ``empathic'' are merged into ``empathy'', whereas generic mentions such as ``round 3'' or ``model 2'' are discarded. 
One author then manually checks the filtered results against the original comments and refines them when necessary to ensure quality.

The resulting word cloud reveals several desired characteristics of model responses. 
Participants consistently value responses that are \textit{helpful}, \textit{actionable}, \textit{clear}, and \textit{supportive}. 
Many also highlight qualities such as \textit{empathy} and \textit{personalization}, suggesting that users care not only about receiving useful guidance but also about feeling understood during the interaction. 
At the same time, the comments reveal substantial diversity in user preferences. 
While some participants appreciate \textit{empathetic validation} before receiving advice, others find such responses overly \textit{long} or too \textit{generic}, and instead prefer more \textit{direct} and \textit{structured} outputs, such as concise actionable to-do lists. 
This diversity helps explain why overall model ranking differences remain small in Table~\ref{tab:overall_ranking}: users do not share a single unified preference over what constitutes a good response.

More importantly, these comments suggest several sub-dimensions that warrant further investigation in ToM-oriented HAI evaluation. 
First, \textbf{belief and knowledge tracking} remains insufficient: participants explicitly criticize models for failing to recognize their lack of context, noting that they ``just jump into solutions'' and should instead ``ask questions like a computer tech would.'' 
Users are also frustrated when models repeat generic advice that they already know or have already tried. 
Second, \textbf{emotional validation} strongly shapes user preference: responses are ranked lower when they sound ``robotic,'' ``uncaring,'' or overly ``clinical,'' whereas top-ranked responses are often praised for making users feel ``heard'' and ``understood.'' 
Third, \textbf{intent recognition} is often mismatched with the user's actual conversational needs. 
For example, several participants feel that strongly suggesting professional help or crisis lines merely because they mention feeling ``down'' is socially inappropriate or ``overkill.'' 
Taken together, these findings highlight that ToM in real interactions goes beyond attributing beliefs and intentions; it also requires adapting to diverse expectations regarding tone, pacing, and response style.

\paragraph{Takeaway}
The user study demonstrates that ToM improvements are perceptible and valuable but also mediated by \textbf{detailed user goals} and \textbf{preferences}. Static benchmarks alone are insufficient; genuine ToM competence in LLMs emerges only when models can flexibly infer intent, balance belief reasoning with pragmatic support, and adapt to heterogeneous human needs.

\subsection{Power Analysis for User Study Setup}
\label{app:us}
\revise{
To assess whether the user study was sufficiently powered, we conducted an \textit{a priori} power analysis based on the non-central chi-square distribution, following the Friedman test and Kendall's \(W\) framework. We set the target effect size to \(W=0.10\), corresponding to weak but perceivable agreement, and the target power to \(0.80\), a commonly adopted standard in behavioral and empirical research~\cite{power}. Under these settings, the required minimum sample size is approximately \(n=37\) for the GPT group (\(k=4\)) and \(n=30\) for the LLaMA group (\(k=5\)). In our study, each group includes \(n=50\) participants, for a total of \(N=100\), indicating that the study is adequately powered to detect systematic human preference if such an effect exists at this level. The absence of strong and consistent preferences in our results therefore suggests that the perceptual differences among current ToM methods, while potentially present, are likely subtle and difficult to capture reliably across participants. This observation further indicates that existing ToM methods may still have limited ability to produce robust and consistently perceivable improvements, motivating the development of more realistic evaluation settings and new ToM approaches. Our sample size is also larger than what is commonly reported in prior human-subject studies. For example, a previous study reports that in user studies published at top-tier venues such as CHI, the median sample size is 20 and the 75th percentile is 46. With \(N=100\) participants overall and 50 participants in each model group, our study exceeds these common research practices~\cite{CHI}.}

% \paragraph{Overall Conclusion.}
% Results show that interaction-based evaluation reveals meaningful improvements in ToM capabilities. Across both open- and closed-source models, the baseline consistently ranked lowest in overall user preference, confirming that static, third-person benchmarks underestimate the practical value of ToM in human–AI interaction.

\section{Additional Cases}\label{app:case}
In this section, we provide more cases to demonstrate the performance of different model variants.
\subsection{Benchmarking Cases}
\begin{figure*}[h]
    \centering
    \resizebox{0.7\textwidth}{!}{%
    \includegraphics{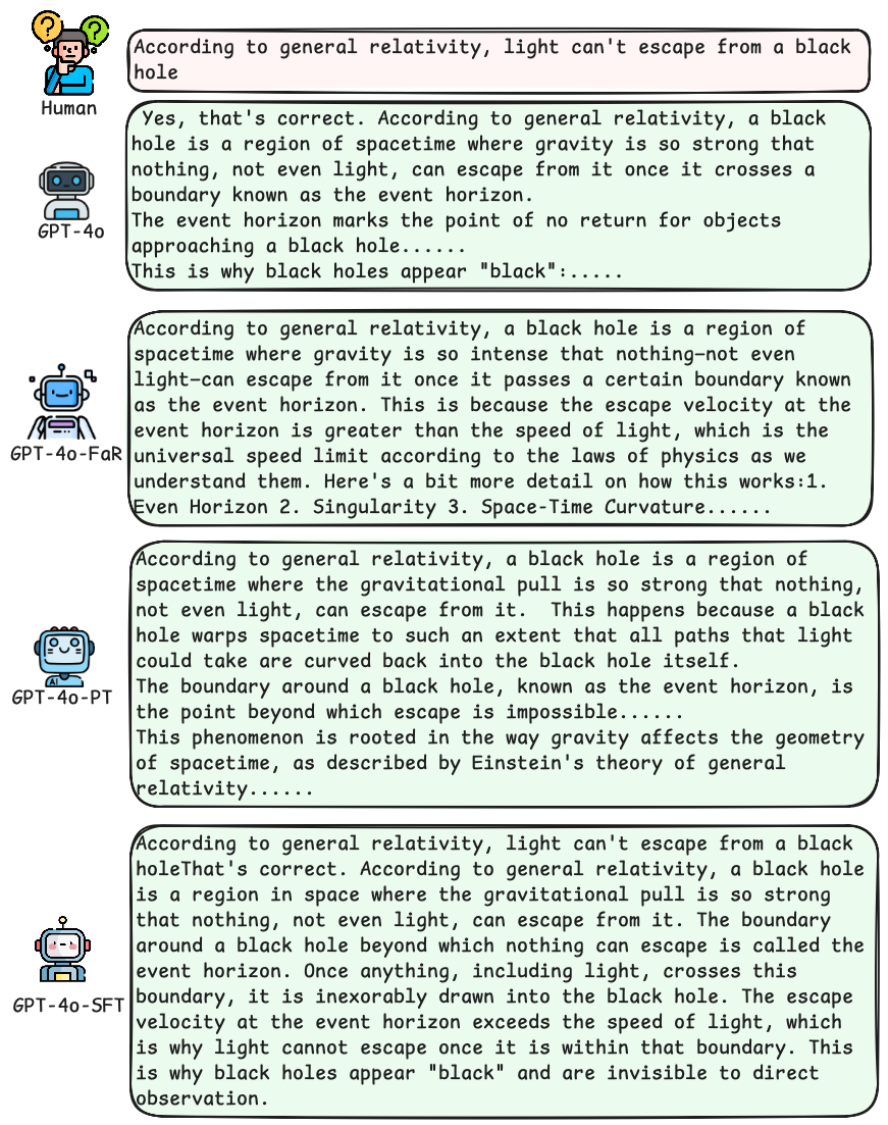}%
    }
    \caption{The case on GPT-4o variants.}
    \label{fig:caseapp1}
    \vspace{-0.4cm}
\end{figure*}

\begin{figure*}[h]
    \centering
    \resizebox{0.7\textwidth}{!}{%
    \includegraphics{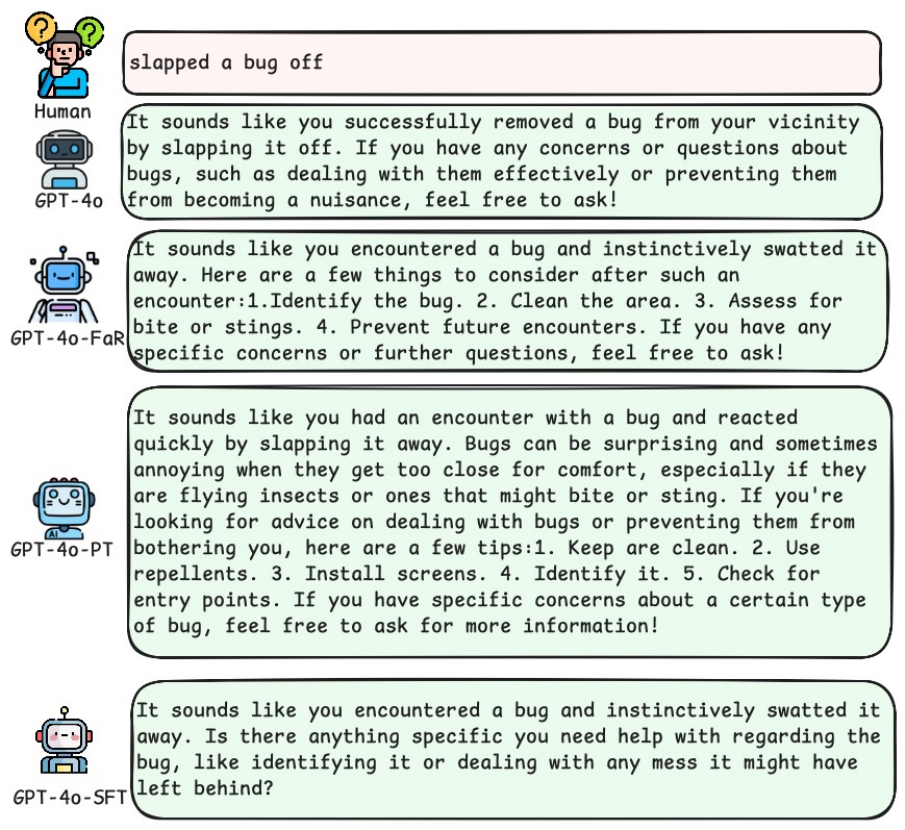}%
    }
    \caption{The case on GPT-4o variants.}
    \label{fig:caseapp2}
    \vspace{-0.4cm}
\end{figure*}

\begin{figure*}[h]
    \centering
    \resizebox{0.7\textwidth}{!}{%
    \includegraphics{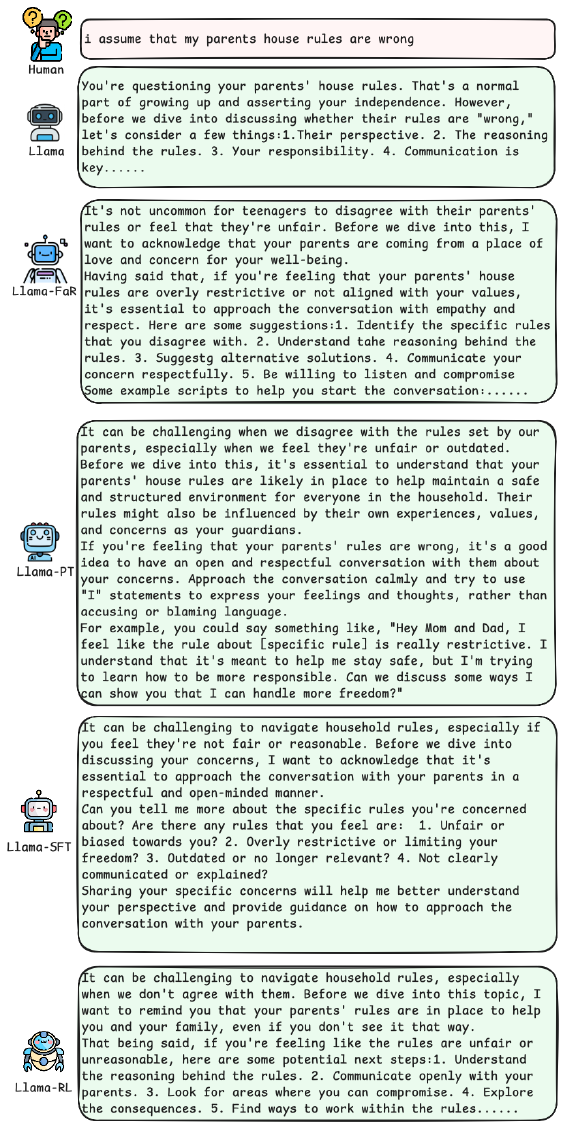}%
    }
    \caption{The case on Llama-3.1-8B variants.}
    \label{fig:caseapp3}
    \vspace{-0.4cm}
\end{figure*}

\begin{figure*}[h]
    \centering
    \resizebox{0.7\textwidth}{!}{%
    \includegraphics{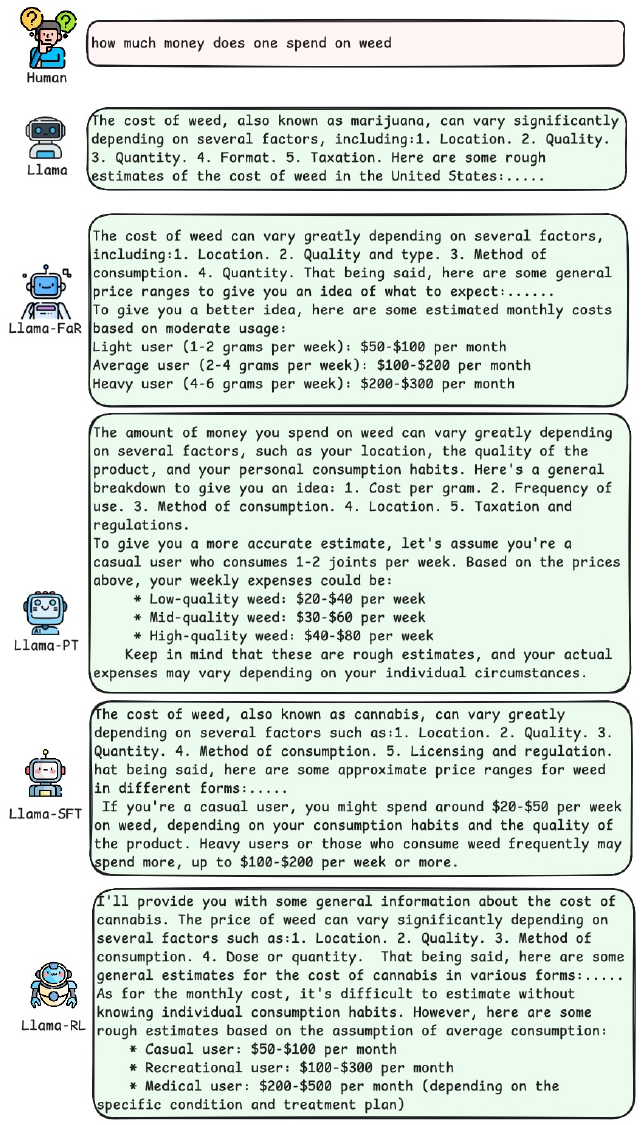}%
    }
    \caption{The case on Llama-3.1-8B variants.}
    \label{fig:caseapp4}
    \vspace{-0.4cm}
\end{figure*}

We provide the cases in our benchmarking process in Figures \ref{fig:caseapp1}-\ref{fig:caseapp4}.


\FloatBarrier
\subsection{User Cases}
\begin{figure*}[h]
    \centering
    \resizebox{0.6\textwidth}{!}{%
    \includegraphics{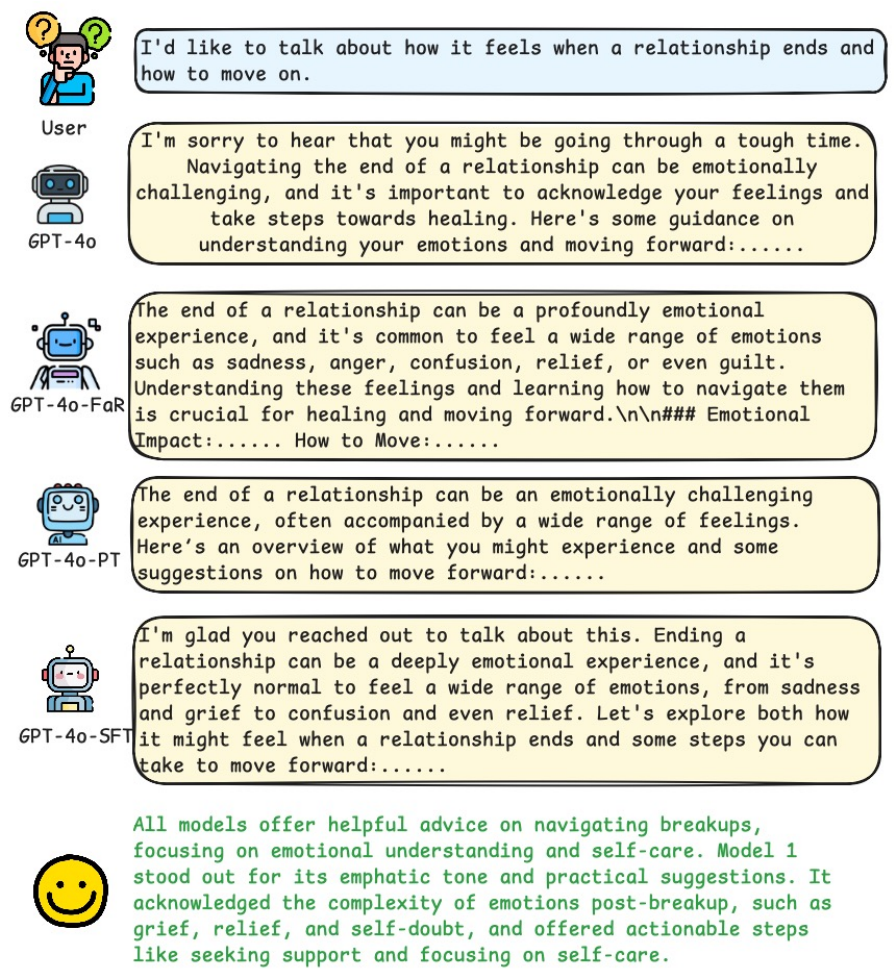}%
    }
    \caption{The user case on GPT-4o variants.}
    \label{fig:userapp1}
    \vspace{-0.4cm}
\end{figure*}

\begin{figure*}[h]
    \centering
    \resizebox{0.6\textwidth}{!}{%
    \includegraphics{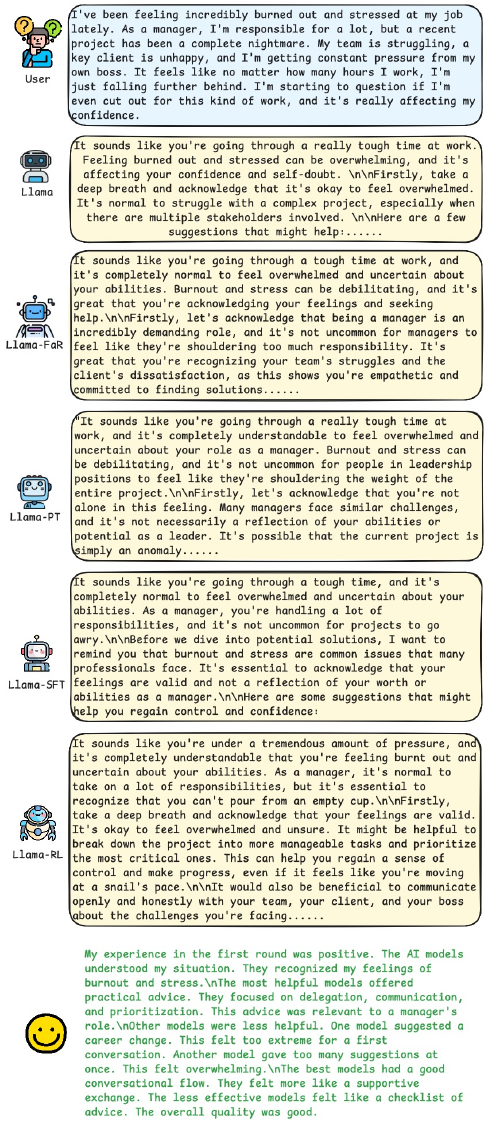}%
    }
    \caption{The user case on Llama-3.1-8B variants.}
    \label{fig:userapp2}
    \vspace{-0.4cm}
\end{figure*}

We provide the cases with participants' comments in our case study in Figures \ref{fig:userapp1}-\ref{fig:userapp2}.

\

\section{LLM Usage Statement}
We used LLMs (e.g., ChatGPT) mainly for grammar and wording edits. Besides, LLMs were used to analyze user study comments to extract keywords related to user metrics.

%% file: Reference.bib
@article{street2024llm,
  title={Llm theory of mind and alignment: Opportunities and risks},
  author={Street, Winnie},
  journal={arXiv preprint arXiv:2405.08154},
  year={2024}
}

@article{strachan2024testing,
  title={Testing theory of mind in large language models and humans},
  author={Strachan, James WA and Albergo, Dalila and Borghini, Giulia and Pansardi, Oriana and Scaliti, Eugenio and Gupta, Saurabh and Saxena, Krati and Rufo, Alessandro and Panzeri, Stefano and Manzi, Guido and others},
  journal={Nature Human Behaviour},
  volume={8},
  number={7},
  pages={1285--1295},
  year={2024},
  publisher={Nature Publishing Group UK London}
}

@article{kosinski2024evaluating,
  title={Evaluating large language models in theory of mind tasks},
  author={Kosinski, Michal},
  journal={Proceedings of the National Academy of Sciences},
  volume={121},
  number={45},
  pages={e2405460121},
  year={2024},
  publisher={National Academy of Sciences}
}

@inproceedings{ToMi,
  title={Revisiting the evaluation of theory of mind through question answering},
  author={Le, Matthew and Boureau, Y-Lan and Nickel, Maximilian},
  booktitle={Proceedings of the 2019 Conference on Empirical Methods in Natural Language Processing and the 9th International Joint Conference on Natural Language Processing (EMNLP-IJCNLP)},
  pages={5872--5877},
  year={2019}
}

@inproceedings{HiToM,
  title={Hi-ToM: A Benchmark for Evaluating Higher-Order Theory of Mind Reasoning in Large Language Models},
  author={Wu, Yufan and He, Yinghui and Jia, Yilin and Mihalcea, Rada and Chen, Yulong and Deng, Naihao},
  booktitle={The 2023 Conference on Empirical Methods in Natural Language Processing},
  year={2023}
}

@inproceedings{ToMchallenges,
  title={ToMChallenges: A Principle-Guided Dataset and Diverse Evaluation Tasks for Exploring Theory of Mind},
  author={Ma, Xiaomeng and Gao, Lingyu and Xu, Qihui},
  booktitle={Proceedings of the 27th Conference on Computational Natural Language Learning (CoNLL)},
  pages={15--26},
  year={2023}
}

@inproceedings{FANToM,
  title={FANToM: A benchmark for stress-testing machine theory of mind in interactions},
  author={Kim, Hyunwoo and Sclar, Melanie and Zhou, Xuhui and Bras, Ronan and Kim, Gunhee and Choi, Yejin and Sap, Maarten},
  booktitle={Proceedings of the 2023 Conference on Empirical Methods in Natural Language Processing},
  pages={14397--14413},
  year={2023}
}

@article{BigToM,
  title={Understanding social reasoning in language models with language models},
  author={Gandhi, Kanishk and Fr{\"a}nken, Jan-Philipp and Gerstenberg, Tobias and Goodman, Noah},
  journal={Advances in Neural Information Processing Systems},
  volume={36},
  pages={13518--13529},
  year={2023}
}

@inproceedings{OpenToM,
  title={OpenToM: A Comprehensive Benchmark for Evaluating Theory-of-Mind Reasoning Capabilities of Large Language Models},
  author={Xu, Hainiu and He, Yulan and Zhu, Lixing and Zhao, Runcong and Du, Jinhua},
  booktitle={The 62nd Annual Meeting of the Association for Computational Linguistics},
  year={2024},
  organization={Association for Computational Linguistics (ACL)}
}

@inproceedings{NegotiationToM,
  title={NegotiationToM: A Benchmark for Stress-testing Machine Theory of Mind on Negotiation Surrounding},
  author={Chan, Chunkit and Jiayang, Cheng and Yim, Yauwai and Deng, Zheye and Fan, Wei and Li, Haoran and Liu, Xin and Zhang, Hongming and Wang, Weiqi and Song, Yangqiu},
  booktitle={Findings of the Association for Computational Linguistics: EMNLP 2024},
  pages={4211--4241},
  year={2024}
}

@inproceedings{ToMBench,
  title={ToMBench: Benchmarking Theory of Mind in Large Language Models},
  author={Chen, Zhuang and Wu, Jincenzi and Zhou, Jinfeng and Wen, Bosi and Bi, Guanqun and Jiang, Gongyao and Cao, Yaru and Hu, Mengting and Lai, Yunghwei and Xiong, Zexuan and others},
  booktitle={Proceedings of the 62nd Annual Meeting of the Association for Computational Linguistics (Volume 1: Long Papers)},
  pages={15959--15983},
  year={2024}
}

@inproceedings{ExploreToM,
  title={Explore Theory of Mind: program-guided adversarial data generation for theory of mind reasoning},
  author={Sclar, Melanie and Dwivedi-Yu, Jane and Fazel-Zarandi, Maryam and Tsvetkov, Yulia and Bisk, Yonatan and Choi, Yejin and Celikyilmaz, Asli},
  booktitle={The Thirteenth International Conference on Learning Representations},
  year = {2025}
}

@inproceedings{ToMATO,
  title={Tomato: Verbalizing the mental states of role-playing llms for benchmarking theory of mind},
  author={Shinoda, Kazutoshi and Hojo, Nobukatsu and Nishida, Kyosuke and Mizuno, Saki and Suzuki, Keita and Masumura, Ryo and Sugiyama, Hiroaki and Saito, Kuniko},
  booktitle={Proceedings of the AAAI Conference on Artificial Intelligence},
  volume={39},
  number={2},
  pages={1520--1528},
  year={2025}
}

@inproceedings{DWM,
  title={A notion of complexity for theory of mind via discrete world models},
  author={Huang, X Angelo and La Malfa, Emanuele and Marro, Samuele and Asperti, Andrea and Cohn, Anthony G and Wooldridge, Michael J},
  booktitle={Findings of the Association for Computational Linguistics: EMNLP 2024},
  pages={2964--2983},
  year={2024}
}

@inproceedings{MP,
  title={Metacognitive prompting improves understanding in large language models},
  author={Wang, Yuqing and Zhao, Yun},
  booktitle={Proceedings of the 2024 Conference of the North American Chapter of the Association for Computational Linguistics: Human Language Technologies (Volume 1: Long Papers)},
  pages={1914--1926},
  year={2024}
}

@inproceedings{PrecepToM,
  title={Perceptions to Beliefs: Exploring Precursory Inferences for Theory of Mind in Large Language Models},
  author={Jung, Chani and Kim, Dongkwan and Jin, Jiho and Kim, Jiseon and Seonwoo, Yeon and Choi, Yejin and Oh, Alice and Kim, Hyunwoo},
  booktitle={Proceedings of the 2024 Conference on Empirical Methods in Natural Language Processing},
  pages={19794--19809},
  year={2024}
}

@inproceedings{TimeToM,
  title={TimeToM: Temporal Space is the Key to Unlocking the Door of Large Language Models’ Theory-of-Mind},
  author={Hou, Guiyang and Zhang, Wenqi and Shen, Yongliang and Wu, Linjuan and Lu, Weiming},
  booktitle={Findings of the Association for Computational Linguistics: ACL 2024},
  pages={11532--11547},
  year={2024}
}

@article{FaR,
  title={How far are large language models from agents with theory-of-mind?},
  author={Zhou, Pei and Madaan, Aman and Potharaju, Srividya Pranavi and Gupta, Aditya and McKee, Kevin R and Holtzman, Ari and Pujara, Jay and Ren, Xiang and Mishra, Swaroop and Nematzadeh, Aida and Upadhyay, Shyam and Faruqui, Manaal},
  journal={arXiv preprint arXiv:2310.03051},
  year={2023}
}

@article{ToM-RL,
  title={Do Theory of Mind Benchmarks Need Explicit Human-like Reasoning in Language Models?},
  author={Lu, Yi-Long and Zhang, Chunhui and Song, Jiajun and Fan, Lifeng and Wang, Wei},
  journal={arXiv preprint arXiv:2504.01698},
  year={2025}
}

@inproceedings{VToM,
  author       = {Zhanwen Chen and
                  Tianchun Wang and
                  Yizhou Wang and
                  Michal Kosinski and
                  Xiang Zhang and
                  Yun Fu and
                  Sheng Li},
  title        = {Through the Theory of Mind's Eye: Reading Minds with Multimodal
                  Video Large Language Models},
  booktitle    = {Proceedings of the International Joint Conference on Neural Networks},
  pages        = {1--8},
  year         = {2025}
}

@inproceedings{Coke,
  title={Coke: A cognitive knowledge graph for machine theory of mind},
  author={Wu, Jincenzi and Chen, Zhuang and Deng, Jiawen and Sabour, Sahand and Meng, Helen and Huang, Minlie},
  booktitle={Proceedings of the 62nd Annual Meeting of the Association for Computational Linguistics (Volume 1: Long Papers)},
  pages={15984--16007},
  year={2024}
}

@article{Thought-Tracing,
  title={Hypothesis-driven theory-of-mind reasoning for large language models},
  author={Kim, Hyunwoo and Sclar, Melanie and Zhi-Xuan, Tan and Ying, Lance and Levine, Sydney and Liu, Yang and Tenenbaum, Joshua B and Choi, Yejin},
  journal={arXiv preprint arXiv:2502.11881},
  year={2025}
}

@inproceedings{AutoToM,
  title={Autotom: Automated bayesian inverse planning and model discovery for open-ended theory of mind},
  author={Zhang, Zhining and Jin, Chuanyang and Jia, Mung Yao and Shu, Tianmin},
  booktitle={ICLR 2025 Workshop on Foundation Models in the Wild},
  year={2025}
}

@inproceedings{Decect_Thoughts,
  title={I cast detect thoughts: Learning to converse and guide with intents and theory-of-mind in dungeons and dragons},
  author={Zhou, Pei and Zhu, Andrew and Hu, Jennifer and Pujara, Jay and Ren, Xiang and Callison-Burch, Chris and Choi, Yejin and Ammanabrolu, Prithviraj},
  booktitle={Proceedings of the 61st Annual Meeting of the Association for Computational Linguistics (Volume 1: Long Papers)},
  pages={11136--11155},
  year={2023}
}

@inproceedings{DecomposeToM,
  title={Decompose-ToM: Enhancing Theory of Mind Reasoning in Large Language Models through Simulation and Task Decomposition},
  author={Sarangi, Sneheel and Elgarf, Maha and Salam, Hanan},
  booktitle={Proceedings of the 31st International Conference on Computational Linguistics},
  pages={10228--10241},
  year={2025}
}

@inproceedings{survey_assess_enhance,
  author       = {Ruirui Chen and
                  Weifeng Jiang and
                  Chengwei Qin and
                  Cheston Tan},
  editor       = {Wanxiang Che and
                  Joyce Nabende and
                  Ekaterina Shutova and
                  Mohammad Taher Pilehvar},
  title        = {Theory of Mind in Large Language Models: Assessment and Enhancement},
  booktitle    = {Proceedings of the 63rd Annual Meeting of the Association for Computational
                  Linguistics (Volume 1: Long Papers)},
  pages        = {31539--31558},
  year         = {2025}
}

@article{survey1,
  title={A systematic review on the evaluation of large language models in theory of mind tasks},
  author={Sar{\i}ta{\c{s}}, Karahan and Tez{\"o}ren, K{\i}van{\c{c}} and Durmazkeser, Yavuz},
  journal={arXiv preprint arXiv:2502.08796},
  year={2025}
}

@article{survey2,
  title={A survey of theory of mind in large language models: Evaluations, representations, and safety risks},
  author={Nguyen, Hieu Minh},
  journal={arXiv preprint arXiv:2502.06470},
  year={2025}
}

@inproceedings{Chatbench,
  author       = {Serina Chang and
                  Ashton Anderson and
                  Jake M. Hofman},
  editor       = {Wanxiang Che and
                  Joyce Nabende and
                  Ekaterina Shutova and
                  Mohammad Taher Pilehvar},
  title        = {ChatBench: From Static Benchmarks to Human-AI Evaluation},
  booktitle    = {Proceedings of the 63rd Annual Meeting of the Association for Computational
                  Linguistics (Volume 1: Long Papers)},
  pages        = {26009--26038},
  year         = {2025}
}

@inproceedings{Collabllm,
  title={CollabLLM: From Passive Responders to Active Collaborators},
year={2025},
  author={Wu, Shirley and Galley, Michel and Peng, Baolin and Cheng, Hao and Li, Gavin and Dou, Yao and Cai, Weixin and Zou, James and Leskovec, Jure and Gao, Jianfeng},
  booktitle={Forty-second International Conference on Machine Learning}
}

@inproceedings{Mentalchat16k,
  title={Mentalchat16k: A benchmark dataset for conversational mental health assistance},
  author={Xu, Jia and Wei, Tianyi and Hou, Bojian and Orzechowski, Patryk and Yang, Shu and Jin, Ruochen and Paulbeck, Rachael and Wagenaar, Joost and Demiris, George and Shen, Li},
  booktitle={Proceedings of the 31st ACM SIGKDD Conference on Knowledge Discovery and Data Mining V. 2},
  pages={5367--5378},
  year={2025}
}

@inproceedings{ESC,
  title={Towards Emotional Support Dialog Systems},
  author={Liu, Siyang and Zheng, Chujie and Demasi, Orianna and Sabour, Sahand and Li, Yu and Yu, Zhou and Jiang, Yong and Huang, Minlie},
  booktitle={Proceedings of the 59th Annual Meeting of the Association for Computational Linguistics and the 11th International Joint Conference on Natural Language Processing (Volume 1: Long Papers)},
  pages={3469--3483},
  year={2021}
}

@article{goal-experience2,
  title={Goals and the Structure of Experience},
  author={Amir, Nadav and Tiomkin, Stas and Langdon, Angela},
  journal={arXiv preprint arXiv:2508.15013},
  year={2025}
}

@incollection{goal-experience1,
  title={Cognitive-experiential self-theory},
  author={Epstein, Seymour},
  booktitle={Advanced personality},
  pages={211--238},
  year={1998},
  publisher={Springer}
}

@article{imuta2016theory,
  title={Theory of mind and prosocial behavior in childhood: A meta-analytic review.},
  author={Imuta, Kana and Henry, Julie D and Slaughter, Virginia and Selcuk, Bilge and Ruffman, Ted},
  journal={Developmental psychology},
  volume={52},
  number={8},
  pages={1192},
  year={2016},
  publisher={American Psychological Association}
}

@article{devine2016theory,
  title={Theory of mind in middle childhood: Longitudinal associations with executive function and social competence.},
  author={Devine, Rory T and White, Naomi and Ensor, Rosie and Hughes, Claire},
  journal={Developmental psychology},
  volume={52},
  number={5},
  pages={758},
  year={2016},
  publisher={American Psychological Association}
}

@article{baron1985does,
  title={Does the autistic child have a “theory of mind”?},
  author={Baron-Cohen, Simon and Leslie, Alan M and Frith, Uta},
  journal={Cognition},
  volume={21},
  number={1},
  pages={37--46},
  year={1985},
  publisher={Elsevier}
}

@misc{Prolific, title={Prolific · Quickly find research participants you can trust.}, url={https://www.prolific.com/}, journal={www.prolific.com}, author={Prolific}, year={2023} }

@article{bales1950interaction,
  title={Interaction process analysis; a method for the study of small groups.},
  author={Bales, Robert F},
  year={1950},
  publisher={Addison-Wesley}
}

@article{liao2023rethinking,
  title={Rethinking model evaluation as narrowing the socio-technical gap},
  author={Liao, Q Vera and Xiao, Ziang},
  journal={arXiv preprint arXiv:2306.03100},
  year={2023}
}

@article{goal_2,
  title={Reading the mind in the eyes or reading between the lines? Theory of mind predicts collective intelligence equally well online and face-to-face},
  author={Engel, David and Woolley, Anita Williams and Jing, Lisa X and Chabris, Christopher F and Malone, Thomas W},
  journal={PloS one},
  volume={9},
  number={12},
  pages={e115212},
  year={2014},
  publisher={Public Library of Science San Francisco, USA}
}

@article{goal_3,
  title={Evidence for a collective intelligence factor in the performance of human groups},
  author={Woolley, Anita Williams and Chabris, Christopher F and Pentland, Alex and Hashmi, Nada and Malone, Thomas W},
  journal={science},
  volume={330},
  number={6004},
  pages={686--688},
  year={2010},
  publisher={American Association for the Advancement of Science}
}

@incollection{goal_1,
  title={Theory of Mind in Human-AI Interaction and AI},
  author={Walsh, Sarah and Wang, Qiaosi and Ying, Lance},
  booktitle={Handbook of Human-Centered Artificial Intelligence},
  pages={1--43},
  year={2025},
  publisher={Springer}
}

@article{experience1,
  title={Large language models as theory of mind aware generative agents with counterfactual reflection},
  author={Yang, Bo and Guo, Jiaxian and Iwasawa, Yusuke and Matsuo, Yutaka},
  journal={arXiv preprint arXiv:2501.15355},
  year={2025}
}

@article{experience2,
  title={Infusing Theory of Mind into Socially Intelligent LLM Agents},
  author={Hwang, EunJeong and Yin, Yuwei and Carenini, Giuseppe and West, Peter and Shwartz, Vered},
  journal={arXiv preprint arXiv:2509.22887},
  year={2025}
}

@article{power,
  title={A power primer.},
  author={Cohen, Jacob},
  year={2016},
  publisher={American Psychological Association}
}

@inproceedings{CHI,
  title={Local standards for sample size at CHI},
  author={Caine, Kelly},
  booktitle={Proceedings of the 2016 CHI conference on human factors in computing systems},
  pages={981--992},
  year={2016}
}
